\documentclass[10pt,twocolumn,letterpaper]{article}

\usepackage[pagenumbers]{cvpr} % To force page numbers, e.g. for an arXiv version

\definecolor{cvprblue}{rgb}{0.21,0.49,0.74}
\usepackage[pagebackref,breaklinks,colorlinks,allcolors=cvprblue]{hyperref}

\def\paperID{*****} % *** Enter the Paper ID here
\def\confName{CVPR}
\def\confYear{2026}

\title{Revisiting Face Recognition for Monozygotic Twins: The Celeb Twins Test Set}

\author{Michael Zang, Haiyu Wu, Mrinal Sharma, Kevin W. Bowyer\\
Department of Computer Science \& Engineering\\
University of Notre Dame\\
{\tt\small (mzang, hwu, msharma2) @alumni.nd.edu, kwb@nd.edu }
}
\begin{document}
\maketitle
\begin{abstract}
Past literature on face recognition for monozygotic (``identical'') twins points to facial marks and mirror asymmetry as possible directions for improved accuracy of twins recognition.
The Celeb Twins Test Set (CTTS) contains web-scraped image pairs for 80 sets of celebrity twins.
It is the only twins test set with meta-data for twins with distinguishing skin marks and possible mirror asymmetry.
CTTS is organized in the manner of face verification test sets such as LFW, CALFW, CPLFW, CFP-FP, and AgeDB-30.
Current deep CNN matchers can achieve over 76\%  accuracy in classifying CTTS same-person / different-person image pairs.
We show that current matchers do {\bf not} make use of skin marks, or asymmetry, and discuss reasons for this.
Finally, we discuss the feasibility of using generative AI tools such as Grok, ChatGPT and Gemini to create images of imagined monozygotic twins as a means to increase representation of twins in face recognition training sets.
\end{abstract}
    
\section{Introduction}
\label{sec:intro}

Monozygotic (MZ) twins occur when a single 
embryo divides into two. 
The global incidence of MZ twins is estimated at between 3 to 4 per 1,000 births \cite{Medline}.
A 2022 National Institute of Standards and Technology (NIST) report \cite{hanaoka2022nist} shows how challenging MZ twins are for face recognition.
In this report, a threshold for 1-to-1 face matching is selected based on the impostor distribution for non-twins to give a 0.0001 (1-in-10,000) false match rate (FMR).
With this threshold, most algorithms in the report had a 0.98 – 0.99 FMR for one twin being accepted as the other.
The lowest FMR for any algorithm in the report was 0.475, meaning that one twin would be accepted as the other about half of the time.

\begin{figure}
\centering
\includegraphics[width=0.23\textwidth]{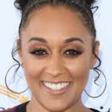}     \includegraphics[width=0.23\textwidth]{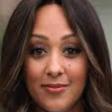}
\\[3pt]
\includegraphics[width=0.23\textwidth]{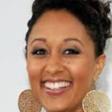}     \includegraphics[width=0.23\textwidth]{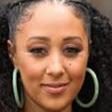}
\\[3pt]
\includegraphics[width=0.23\textwidth]{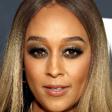}
\includegraphics[width=0.23\textwidth]{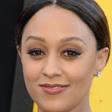}
\caption{
{\it Which row contains an image pair of the two MZ twins rather than two images of the same twin?} Hint: consider freckles.  Images shown are 112x112, as input to CNN matchers.
}
\label{fig:example_image_pairs}
\end{figure}

\begin{figure}
\centering
\includegraphics[width=8.25cm]{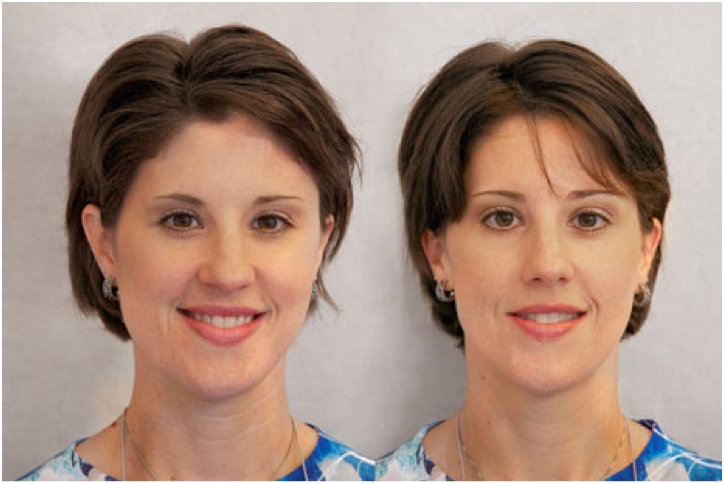}
\caption{Example images from the ND Twins Days dataset.  Note the coordination of hairstyle, cosmetics, clothing, necklace and earrings. This is common in images from the Twins Days Festival.
}
\label{fig:ND_Twins_2009_2010}
\end{figure}

One factor holding back research on automated face recognition for twins is lack of experimental datasets, for both training and testing.
The ND-Twins test set was recently introduced to address this \cite{Wu2026}.
However, image pairs in ND-Twins are drawn from the ND Twins Days dataset \cite{paone2014double}; see Figure \ref{fig:ND_Twins_2009_2010}.
The ND-Twins test set and the larger ND Twins Days dataset do not have meta-data for the presence of distinguishing skin marks\footnote{
There are face image datasets with metadata for skin marks; e.g., \cite{Nexdata, Unidata}. 
But these are not focused on twins.
} or for ``mirror'' twin status.
This paper analyzes the Celeb Twins Test Set, CTTS-80, which contains 264 image pairs for each of 80 sets of celebrity twins, for a total of 21,120 image pairs.
This is over 3.5 times the 6,000 image pairs in ND-twins \cite{Wu2026}, LFW \cite{lfw}, CALFW \cite{calfw}, CPLFW \cite{cplfw}, CFP-FP \cite{cfpfp}, AgeDB-30 \cite{agedb-30}, Hadrian \cite{Wu2026} and Eclipse \cite{Wu2026}.
Over half the twins in CTTS-80 have skin marks that can distinguish between the twins; see Figures \ref{fig:example_image_pairs}, \ref{fig:Mowry_twins}, \ref{fig:Dobre_twins}, \ref{fig:Hassan_twins} and \ref{fig:Kaczynski_twins}.
In addition, 7 sets of twins are identified as one twin being left-handed and the other right-handed, evidence of them being mirror twins.

The remainder of this paper is organized as follows.
Section \ref{sec:twin_background} reviews selected elements of MZ twin biology.
Section \ref{sec:human_ability} reviews research on human ability to distinguish MZ twins.
Section \ref{sec:afr} reviews research on automated face recognition for MZ twins.
Section \ref{sec:ctts} describes CTTS and presents accuracy of face recognition algorithms on CTTS.
Section 6 analyzes whether current matchers make use of visible skin marks and mirror asymmetry.
Section 7 looks at whether current commercial Gen AI tools can produce images of non-existent twins that could be used as training data to create matchers with better accuracy for twins. 
Lastly, Section 8 discusses conclusions and future research directions.

\section{MZ Twins: Biological Background}
\label{sec:twin_background}

The default view of MZ twin biology is that a single embryo divides, resulting in two embryos with ``the same'' DNA.
The default view of this for face recognition is that the two persons have ``the same'' facial appearance.
It should not be surprising that the default view is overly simplistic.

The term ``mirror twins’’ refers to a particular subset of MZ twins, where the ``handedness’’ of some physical features is reversed between the twins \cite{Bingyul2025, Zapata2020}.  For instance, MZ twins who are mirror twins may have one twin be left-handed and the other right-handed, or may have a hair whorl that is clockwise for one twin and counter-clockwise for the other, or reversed left-right dental patterns, among other possibilities.  What exactly is mirrored varies between sets of mirror twins.  A particular set of mirror twins may or may not have an asymmetry in facial appearance that is useful in distinguishing between them.

Most MZ twins occur from the embryo splitting in the range of around four days after conception \cite{Bingyul2025}.  For the mirror subset of MZ twins the embryo splits later, in the range of about 7 to 10 days after conception, when the embryo has begun to develop a left and a right side \cite{Bingyul2025}.  There is no DNA test to determine whether or not MZ twins are mirror twins.  It is estimated that about 1 in 4 MZ twins are mirror twins.  

It is sometimes said that moles or freckles are an element of mirror twin appearance; e.g., see \cite{Kirale2020, Zapata2020, Fierro2024}. However, there appears to be no basis for this.  Mirror twins may or may not have birth marks in mirrored locations, and generally do not have skin marks in mirrored locations. And, non-mirror MZ twins generally do not have skin marks in the same location. Skin marks such as moles and freckles are one source of information to distinguish between MZ twins generally rather than only mirror twins.

A simple way to detect some mirror twins is that one is left-handed and the other right-handed. But not all mirror twins are opposite-handed.  One study suggests that something in the range of 50\% to 75\% of mirror twins are opposite-handed \cite{MultiplesSurvey1987}.  We use known opposite-handedness to categorize some of the MZ twins in CTTS as mirror twins.  It is likely that some CTTS twins not categorized as mirror twins are also mirror twins.

Facial appearance naturally changes over time.
Aging effects may differ between MZ twins due to environmental conditions, health habits, or other factors.
The practical impact is that an older pair of MZ twins has an increased chance of differences that make them easier to distinguish.
The NIST study mentioned earlier \cite{hanaoka2022nist} noted lower average similarity between MZ twins’ faces with increased age.

Age also impacts DNA.  
Even if the DNA of the two embryos is identical at the moment the embryo divides, future accumulated mutations can be enough to distinguish between MZ twins based on their DNA. 
This is done through ``next-generation sequencing’’ or ``deep whole genome sequencing’’ \cite{Kay2021}.
However, there is no reason to believe that mutations that allow distinguishing between MZ twin DNA cause any change in facial appearance. 

In summary, while MZ twins may be called ``identical’’, this is not to be taken literally.  
MZ twins generally have some small differences in facial appearance, and the presence of such differences tend to increase with age.
One element of this is skin marks such as moles, freckles or scars present on one twin but not the other.
Another possible element is mirrored asymmetry in some aspect of face appearance.

\section{Human Ability to Distinguish Twins}
\label{sec:human_ability}

Biswas \etal \cite{biswas2011study} report on experiments that use images from the ND Twins Days dataset \cite{paone2014double}.  
(See Figure~\ref{fig:ND_Twins_2009_2010}.)
Images are frontal pose, controlled acquisition, and cropped to the face region based on eye locations.  
In one experiment, 25 subjects viewed 100 image pairs, with unlimited viewing time, to rate image pairs on a 5-point scale for whether they believe the images are of the same person or of MZ twins.  Mean accuracy was 93\%, with a high of 100\% and a low of 78\%.  Subjects in this experiment were also asked what features supported their decision on each image pair.  The response ``moles / scars / freckles'' was selected on over 40\% of correctly classified image pairs and only 15\% of incorrectly classified pairs.  In contrast, ``eyes'', ``nose'' and ``lips'' were selected for over 15\% of incorrectly classified  pairs, and for less than 10\% of correctly classified  pairs.  

Martini \etal \cite{martini2015me} asked twins to categorize the person in a cropped face image shown for 30 ms as either them, their twin, or a close friend.  Experiments were done with images from 10 sets of MZ twins.  They report that twins are no more accurate at identifying their own image than the image of their twin, and that close friends are no more accurate at identifying one twin than the other.  We speculate that the 30 ms viewing time is too short for twins or friends to reliably employ any strategy based on using local facial features or mirror asymmetry of facial features.

Parde \etal \cite{Parde2023} compare human and algorithm ability at distinguishing twins.  They use image pairs from the ND Twins Days dataset \cite{paone2014double}, with (frontal, frontal), (frontal, $45^{\circ)}$ and (frontal, $90^{\circ)}$ image pairs for same-person, twin impostor and general impostor pairs.  Each of 87 subjects viewed 120 image pairs.
A ResNet101-based matcher \cite{Ankan2017} was also used to compute similarity  for the  image pairs.  They report that ``… DCNN accuracy exceeded the accuracy of nearly all human participants tested in all conditions” \cite{Parde2023}.  Subjects were students; trained face examiners would almost certainly have higher accuracy.  The matcher used is also almost certainly not competitive with  current ResNet-based matchers.

Srinivas \etal \cite{Srinivas2012} studied ``facial marks’’ as a means to distinguish identical twins.  Facial marks were defined broadly to include moles, freckles, light / dark spots, birthmark, scar and other features.  
Manual annotation and automated detection of facial marks are compared, and bipartite graph matching is used for measuring similarity between sets of facial marks.  They report equal error rates of about 24\% - 35\% for twin recognition based on manually annotated facial marks from different annotators and about 27\% for automatically annotated facial marks (see Fig. 27 of \cite{Srinivas2012}).  This is a radically different approach to twin recognition than in current deep CNN face matchers, using much higher resolution images.
ND Twins Days images \cite{paone2014double} have average inter-pupillary distance of 567 pixels.  

Wilmer \etal \cite{wilmer2010heritable} examined whether human face recognition ability is an inherited skill.  They recruited 164 MZ twin pairs and 125 same-gender dizygotic (DZ) pairs and had each person take the Cambridge Face Memory Test (CMFT).  They compared the correlation between CFMT scores for MZ twins, 0.70, versus for DZ twins, 0.29, and infer that the level of face recognition ability is strongly influenced by genetics.
% – ``Our twin findings indicate that genetic differences can account for most of the stable variation in face recognition ability in healthy adults.’’  
This result may suggest that comparing algorithm and human face recognition accuracy should be based on best-performing human rather than the average of a group of humans.

Stevenage \cite{Stevenage1998} investigated whether training improves ability to discriminate twins.  In one experiment, 19 subjects rated similarity of head-and-shoulders images of the same person (e.g., ``Elizabeth-Elizabeth’’) or the twins (``Elizabeth-Rosie’’).  This was followed by a training phase in which subjects labeled images by name, progressing until they reached a 90\% correct.  Subjects then repeated the rating of similarity of image pairs.  After training, same-person image pairs were rated as significantly more similar and twin image pairs as significantly more different.  
A second experiment involving a fixed amount of training produced a similar result.
We speculate that the improvement in rating of similarity of image pairs could result from subjects using the training phase to learn local face features that allow them to distinguish the particular twins.

Lessons from the literature on human ability to distinguish MZ twins include the following.
\begin{itemize}
\item Facial marks, broadly defined, appears to be an important means for humans to accurately distinguish twins \cite{biswas2011study, Srinivas2012}.
\item  We do not find any explicit mention of the use of asymmetry features in the literature on human ability to distinguish twins.
\item Face recognition ability appears to vary substantially across persons \cite{wilmer2010heritable}.
\item While face recognition ability is influenced by genetics, it can be improved by training \cite{wilmer2010heritable, Stevenage1998}.
\end{itemize}

\section{Automated Face Recognition of MZ Twins}
\label{sec:afr}

Research into automated face recognition for twins began around 2010, with a Chinese Academy of Sciences (CASIA) report exploring different biometric modalities for twins \cite{sun2010study}, and results from the ND Twins Days dataset \cite{pruitt2011facial, phillips2011distinguishing} and general availability of that dataset to the research community.  The ND Twins Days dataset has been used in publications by a number of researchers \cite{Klare_2011, biswas2011study, Le_2012, Juefei-Xu_2013, Mahalingam_2013, Mahalingam_2013b, Shalin_2015, afaneh2017fusion, Sudhakar_2021, Parde2023, Brahmbhatt2026}.

Pruitt \etal \cite{pruitt2011facial} reported results for three commercial matchers and a local-region principal components analysis (PCA) algorithm on the ND Twins Days dataset.
Phillips \etal \cite{phillips2011distinguishing} expanded on this, including results for matching images acquired across the 2009 and 2010 Twins Days Festivals.
Paone \etal \cite{paone2014double} expanded further on this with more detailed covariates and additional matchers.

Klare \etal \cite{Klare_2011} explored the use of  different ``levels'' of features.
Level 1 features describe holistic elements of face appearance.
(This work is from before the rise of CNNs for face recognition, but CNN embeddings would likely be level 1 features.)
Level 2 features are things like local binary patterns (LBP), scale-invariant feature transform (SIFT) and Gabor filters.
Level 3 features are facial marks.
To abstract from inaccuracies in automated detection of facial marks, facial marks were manually annotated for this work.
One interesting observation from this work is that facial marks were relatively more important to increasing accuracy for twins than for non-twins.

Le \etal \cite{Le_2012} use a two-step approach to recognize twins. 
The first step uses Local Fisher Discriminant Analysis (LFDA) to classify an image as from a particular set of twins, and the second step uses Gabor filter features from specific face regions to distinguish between the twins.
The features are related to face aging in that, as mentioned earlier, MZ twin faces can acquire unique features as the individuals age.
Juefei-Xu and Savvides \cite{Juefei-Xu_2013} use facial asymmetry features and Augmented Linear
Discriminant Analysis in recognizing MZ twins.
Le \etal \cite{Le_2015} build on the work of \cite{Le_2012} and \cite{Juefei-Xu_2013} to use both asymmetry features and local face region features in recognizing MZ twins.
These works \cite{Juefei-Xu_2013, Le_2015} are the only ones we are aware of to emphasize the use of asymmetry features.
One limitation is that the asymmetry features used assume a controlled frontal pose.

Mahalingam and Ricanek \cite{Mahalingam_2013} analyze accuracy of LBP- and HOG-based algorithms for recognizing twins.  This is the only paper we found other than \cite{sun2010study} that reports results on the CASIA dataset, and it also includes comparison results on the ND Twins Days dataset.  Accuracies for the CASIA dataset are lower, possibly due to a number of the subjects being children. They evaluated accuracy across age groups and concluded that ``… discriminability increases between the identical twins as the age progresses’’ \cite{Mahalingam_2013}.

Hu \etal \cite{Hu_2015} assemble the fine-grained face verification (FGFV) dataset and explore using HOG, LBP and SIFT features to classify an image pair as positive / negative. 
The FGFV dataset is 455 positive image pairs, selected as 2 images of the same person from the LFW dataset, and 455 negative image pairs, selected as images of a pair of twins.
Images of a pair of twins are cropped from the same image sourced from the web; FGFV has only one image of a person in a pair of twins.
This 100\% correlation of image source and category raises the possibility that the classifier learns positive / negative to mean from different / same source image \cite{Lopez_2016, Bowyer_2020}.
Their best combination of classifier and features achieves just over 72\% accuracy in classifying positive / negative pairs.

Shalin \etal \cite{Shalin_2015} argue that automatically detected facial marks are an effective way to distinguish twins, and claim to use the ND Twins Days dataset in experiments.
However, Figure 1 in \cite{Shalin_2015} appears to be a copy of Figure 9 in \cite{Srinivas2012}, and Figure 2 in \cite{Shalin_2015} appears to be a copy of Figure 4 in \cite{paone2014double}. While \cite{Srinivas2012} deals with twins recognition from facial marks, \cite{paone2014double} does not, and \cite{paone2014double} is not cited in \cite{Shalin_2015}.

Mousavi \etal \cite{Mousavi_2021} propose a distinctive landmark-based face recognition (DLFR) approach, using an M-SIFT algorithm to find landmark points on the face.
They also create a dataset of 110 pairs of MZ twin images and 100 pairs of non-twin images.
Some example images in the paper are of babies and children, and some of adults.

Afaneh \etal \cite{afaneh2017fusion} explored twins recognition using fusion at feature-, score- and decision-level  with PCA, HOG and LBP features.
They also report results for an early deep CNN face matcher.
Sudhakar and Nithyanandam \cite{Sudhakar_2021} propose a multi-feature, multi-level fusion approach to distinguishing twins, similar to \cite{afaneh2017fusion},
but using more types of features than \cite{afaneh2017fusion}.
Interestingly, they convert color images to grayscale for further processing.
No deep CNN matchers are evaluated in this work.

Ahmad \etal \cite{Ahmad_2019} report on training an ArcFace-style face matcher with triplet loss for twin recognition.
The experimental dataset apparently consists of 1400 images of each person in 5 sets of twins, for a total of 14,000 images, obtained from Shutterstock.com.
The data is split into 7 parts, with 6 used for training and 1 for validation.
Apparently all twins are represented in each of the 7 parts, and so there is no estimate of separate test accuracy.

Sundaresan and Shanthi \cite{Sundaresan_2021} categorize twins face recognition efforts and suggest topics for future research.
The categorization of primary types of features includes facial marks, facial aging features and facial asymmetry, which is essentially equivalent to facial marks and asymmetry, with either being from birth or aging.
There are no new experimental results in this paper.

Sami \etal \cite{Sami_2022} explore similarity of facial appearance in MZ twins and in unrelated look-alikes or “doppelgangers”.
They use the WV Twins Days dataset, a superset of the ND Twins Days dataset, with additional collections through Twins Days 2018.
(The paper states, ``The data that support the findings of this paper are available from the corresponding author upon reasonable request.'')
Twins discrimination results are shown for a range of deep CNN matchers.
An interesting element of this work is that it develops a metric of facial similarity distinct from an embedding and similarity metric used for recognition.
Look-alikes are defined in terms of similarity relative to a metric derived from a twins dataset, and an interesting conclusion is that “look-alike pairs are quite rare in the populations contained within the data sets used in this study”.

Hsu and Tsai \cite{Hsu_2023} evaluated multiple deep CNN matchers on their own dataset of 182 images of 54 sets of twins.
They fine-tune the pre-trained deep CNN that initially achieves  highest accuracy, using a set of 31 twin pairs.
However, the fine-tuning step does not increase overall accuracy on twins, which is 61\% with and without fine-tuning, and much higher on positive image pairs (77\%) than negative image pairs (45\%).
There is no mention of the twins image dataset being made available to other researchers.

The largest evaluation of face matching for MZ twins is the NIST report \cite{hanaoka2022nist} mentioned earlier.
The evaluation uses a 1-in-10,000 FMR threshold determined from a large set of mugshot-quality, non-twins image pairs.
The fraction of the similarity scores for pairs of images of MZ twins that exceeds the 1-in-10,000 threshold is the twins FMR.
Twins images come from the dataset used in \cite{Sami_2022}.
One conclusion is – ``In the NIST face recognition verification test, algorithms have high false matches and cannot distinguish identical and fraternal same-sex twins as different people.’’
The report also notes – ``Facial landmarks are distinctive and useful, but usually there only few of these features on the face and sometimes they cannot be observed ...’’

Brahmbhatt \etal \cite{Brahmbhatt2026} describe an approach to using automatically detected skin marks to distinguish twins.
They use a pre-trained deep network to detect various types of skin marks, and build a feature vector using the location and type of skin marks.
Using the ND Twins Days dataset, they select the best image of each person for their method.
They end up analyzing 319 images from 74 sets of twins, and report an accuracy of 96\%.

Observations from previous work on automated face recognition include the following.
\begin{itemize}
    \item As with human ability to distinguish twins, facial marks appear to be useful \cite{Klare_2011, Le_2012, Le_2015, Mousavi_2021, Sundaresan_2021, hanaoka2022nist, Brahmbhatt2026}.
    \item Features related to facial asymmetry may be useful but there is limited work in this area \cite{Juefei-Xu_2013, Le_2015}.
    \item The most widely used MZ twins dataset is the ND Twins Days dataset, which is now 15+ years old, and too small for effective deep CNN training.
    \item It is difficult to compare  accuracy across papers, due to lack of a standard face verification test set.
\end{itemize}

\newpage
\begin{figure*}
\centering
\includegraphics[width=0.15\textwidth]{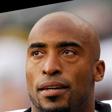}
\includegraphics[width=0.15\textwidth]{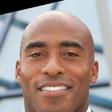}
\includegraphics[width=0.15\textwidth]{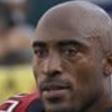}
\includegraphics[width=0.15\textwidth]{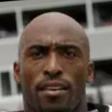}
\includegraphics[width=0.15\textwidth]{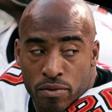}
\includegraphics[width=0.15\textwidth]{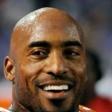}
\\
\includegraphics[width=0.15\textwidth]{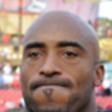}
\includegraphics[width=0.15\textwidth]{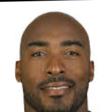}
\includegraphics[width=0.15\textwidth]{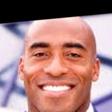}
\includegraphics[width=0.15\textwidth]{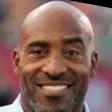}
\includegraphics[width=0.15\textwidth]{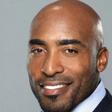}
\includegraphics[width=0.15\textwidth]{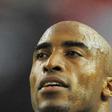}
\\
(a) Twelve 112x112 normalized face crops - Ronde Barber
\\[6pt]
\includegraphics[width=0.15\textwidth]{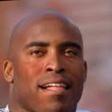}
\includegraphics[width=0.15\textwidth]{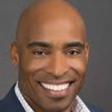}
\includegraphics[width=0.15\textwidth]{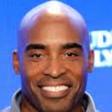}
\includegraphics[width=0.15\textwidth]{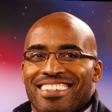}
\includegraphics[width=0.15\textwidth]{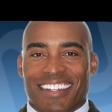}
\includegraphics[width=0.15\textwidth]{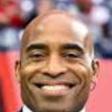}
\\
\includegraphics[width=0.15\textwidth]{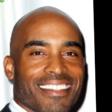}
\includegraphics[width=0.15\textwidth]{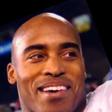}
\includegraphics[width=0.15\textwidth]{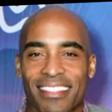}
\includegraphics[width=0.15\textwidth]{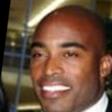}
\includegraphics[width=0.15\textwidth]{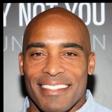}
\includegraphics[width=0.15\textwidth]{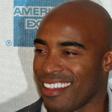}
\\
(b) Twelve 112x112 normalized face crops - Tiki Barber
\\[6pt]
\includegraphics[width=0.85
\textwidth]{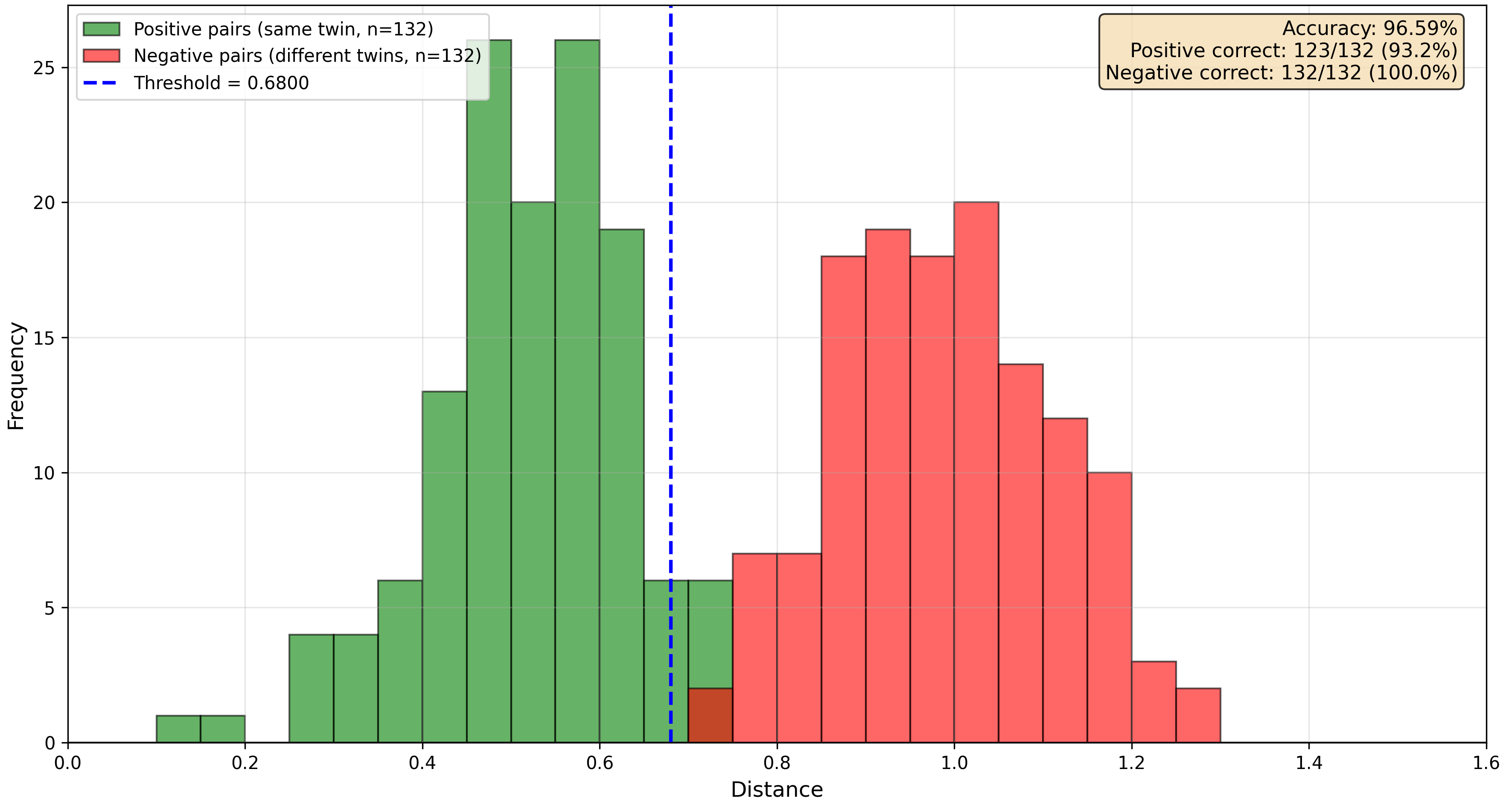}
\\
(c) Distance Distributions of Same-Person (``positive'') and Different-Person (``negative'') Image Pairs
\caption{Example of Well-Separated Distance Score Distributions. 
Euclidean distance of embeddings from ArcFace ResNet-100 backbone trained on MS1MV2.  
Threshold selected from other 9 folds in 10-fold cross-val is lower than ideal for these distributions, but accuracy still exceeds 96\%.  Note image variation in face expression, facial hair, eyeglasses, background and other factors.
}
\label{fig:Barber_twins}
\end{figure*}

\begin{figure*}
\centering
\includegraphics[width=0.15\textwidth]{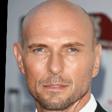}
\includegraphics[width=0.15\textwidth]{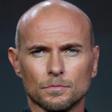}
\includegraphics[width=0.15\textwidth]{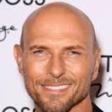}
\includegraphics[width=0.15\textwidth]{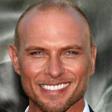}
\includegraphics[width=0.15\textwidth]{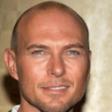}
\includegraphics[width=0.15\textwidth]{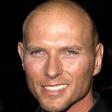}
\\
\includegraphics[width=0.15\textwidth]{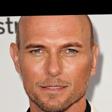}
\includegraphics[width=0.15\textwidth]{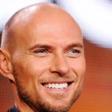}
\includegraphics[width=0.15\textwidth]{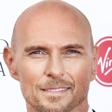}
\includegraphics[width=0.15\textwidth]{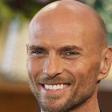}
\includegraphics[width=0.15\textwidth]{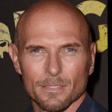}
\includegraphics[width=0.15\textwidth]{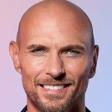}
\\
(a) Twelve 112x112 normalized face crops - Luke Goss
\\[6pt]
\includegraphics[width=0.15\textwidth]{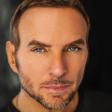}
\includegraphics[width=0.15\textwidth]{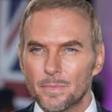}
\includegraphics[width=0.15\textwidth]{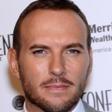}
\includegraphics[width=0.15\textwidth]{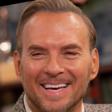}
\includegraphics[width=0.15\textwidth]{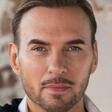}
\includegraphics[width=0.15\textwidth]{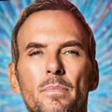}
\\
\includegraphics[width=0.15\textwidth]{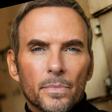}
\includegraphics[width=0.15\textwidth]{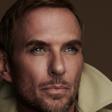}
\includegraphics[width=0.15\textwidth]{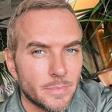}
\includegraphics[width=0.15\textwidth]{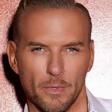}
\includegraphics[width=0.15\textwidth]{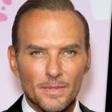}
\includegraphics[width=0.15\textwidth]{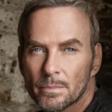}
\\
(b) Twelve 112x112 normalized face crops - Matt Goss
\\[6pt]
\includegraphics[width=0.85
\textwidth]{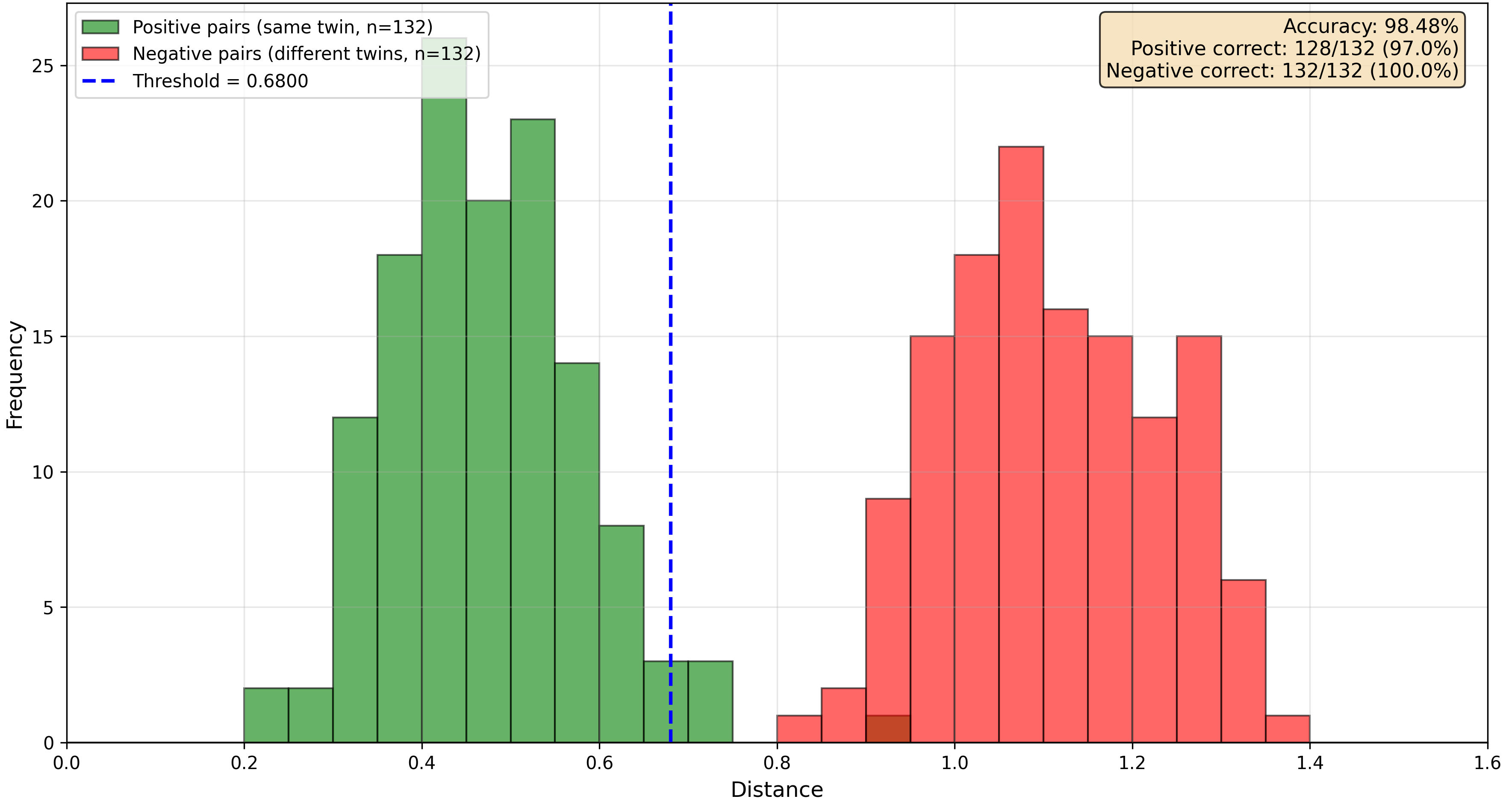}
\\
(c) Distance Distributions of Same-Person (``positive'') and Different-Person (``negative'') Image Pairs
\caption{Example of Well-Separated Distance Score Distributions. Euclidean distance of embeddings from ArcFace ResNet-100 backbone trained on MS1MV2.  
Threshold selected from other 9 folds in 10-fold cross-val is lower than ideal for these distributions.
}
\label{fig:Goss_twins}
\end{figure*}

\begin{figure*}
\centering
\includegraphics[width=0.15\textwidth]{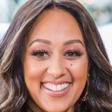}
\includegraphics[width=0.15\textwidth]{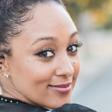}
\includegraphics[width=0.15\textwidth]{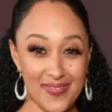}
\includegraphics[width=0.15\textwidth]{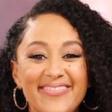}
\includegraphics[width=0.15\textwidth]{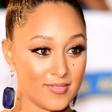}
\includegraphics[width=0.15\textwidth]{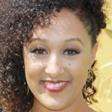}
\\
\includegraphics[width=0.15\textwidth]{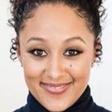}
\includegraphics[width=0.15\textwidth]{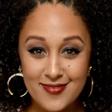}
\includegraphics[width=0.15\textwidth]{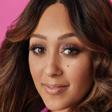}
\includegraphics[width=0.15\textwidth]{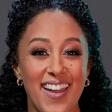}
\includegraphics[width=0.15\textwidth]{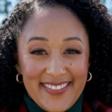}
\includegraphics[width=0.15\textwidth]{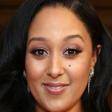}
\\
(a) Twelve 112x112 normalized face crops - Tamera Mowry
\\[6pt]
\includegraphics[width=0.15\textwidth]{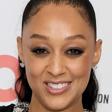}
\includegraphics[width=0.15\textwidth]{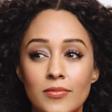}
\includegraphics[width=0.15\textwidth]{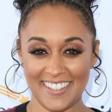}
\includegraphics[width=0.15\textwidth]{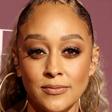}
\includegraphics[width=0.15\textwidth]{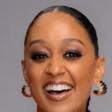}
\includegraphics[width=0.15\textwidth]{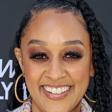}
\\
\includegraphics[width=0.15\textwidth]{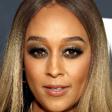}
\includegraphics[width=0.15\textwidth]{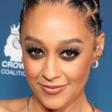}
\includegraphics[width=0.15\textwidth]{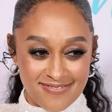}
\includegraphics[width=0.15\textwidth]{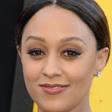}
\includegraphics[width=0.15\textwidth]{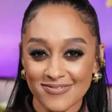}
\includegraphics[width=0.15\textwidth]{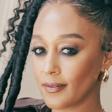}
\\
(b) Twelve 112x112 normalized face crops - Tia Mowry
\\[6pt]
\includegraphics[width=0.85
\textwidth]{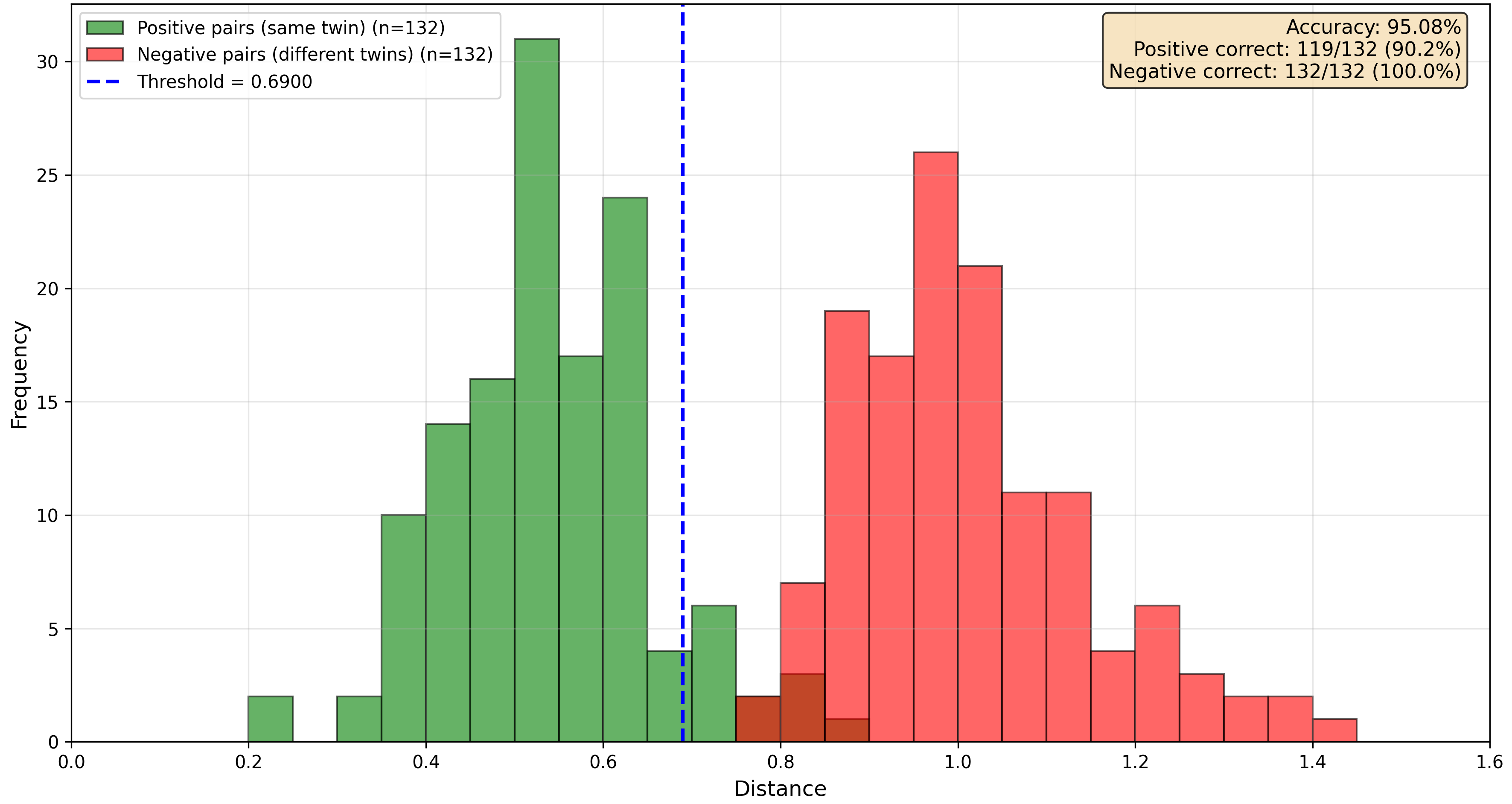}
\\
(c) Distance Distributions of Same-Person (``positive'') and Different-Person (``negative'') Image Pairs
\caption{Example of Well-Separated Distance Score Distributions. 
Euclidean distance of embeddings from ArcFace ResNet-100 backbone trained on MS1MV2.
Threshold selected from other 9 folds in 10-fold cross-val is lower than ideal for these distributions.  Tamera Mowry has a skin mark visible in these images below her left eye that distinguishes  the twins.
}
\label{fig:Mowry_twins}
\end{figure*}

\section{The Celeb Twins Test Set}
\label{sec:ctts}

\begin{table*}
\centering
\renewcommand{\arraystretch}{1.25}
\begin{tabular}{|l|l||c|c|c|c|c|}
\hline
\textbf{Model } & 
\textbf{Train set} &
\textbf{~Trad 5~} &
\textbf{~Hadrian~} &
\textbf{~Eclipse~} &
\textbf{~ND-Twins~} &
\textbf{~CTTS-80~} \\ 
\hline \hline

AdaFace & MS1MV2 & 97.17 & $92.35 \pm 1.85$ & $82.32 \pm 1.66$ & $68.55 \pm 3.01$ & $76.20 \pm 3.73$ \\
AdaFace & WebFace4M & 97.40 & $91.07 \pm 2.19$ & $82.80 \pm 1.63$ & $72.48 \pm 3.05$ & $75.22 \pm 3.69$ \\
AdaFace & glint360k & 97.69 & $95.63 \pm 1.64$ & $83.88 \pm 1.56$ & $75.88 \pm 2.48$ & $75.46 \pm 4.09$ \\
\hline

ArcFace & MS1MV2 & 97.14 & $91.72 \pm 1.78$ & $81.80 \pm 1.19$ & $68.13 \pm 3.24$ & $76.50 \pm 3.42$ \\
ArcFace & WebFace4M & 97.44 & $91.78 \pm 2.14$ & $83.20 \pm 1.62$ & $71.73 \pm 3.08$ & $74.53 \pm 4.07$ \\
ArcFace & glint360k & 97.60 & $95.68 \pm 1.65$ & $84.12 \pm 1.78$ & $84.65 \pm 2.87$ & $74.39 \pm 4.60$ \\
\hline

UniFace & MS1MV2 & 97.13 & $91.63 \pm 2.14$ & $82.18 \pm 1.83$ & $65.75 \pm 2.77$ & $76.16 \pm 4.35$ \\
UniFace & WebFace4M & 97.41 & $90.65 \pm 1.91$ & $82.20 \pm 1.55$ & $71.88 \pm 4.27$ & $75.06 \pm 3.87$ \\
UniFace & glint360k & 97.56 & $93.53 \pm 1.90$ & $83.18 \pm 1.35$ & $71.72 \pm 2.82$ & $74.94 \pm 3.58$ \\
\hline

MagFace & MS1MV2 & 97.00 & $92.82 \pm 2.49$ & $82.67 \pm 1.73$ & $68.57 \pm 2.12$ & $75.71 \pm 3.85$ \\
MagFace & WebFace4M & 97.41 & $90.92 \pm 2.13$ & $82.47 \pm 1.28$ & $71.32 \pm 3.68$ & $73.98 \pm 4.29$ \\
MagFace & glint360k & 97.56 & $95.20 \pm 1.53$ & $83.78 \pm 1.38$ & $71.98 \pm 3.13$ & $73.14 \pm 4.19$ \\
\hline
\end{tabular}
\caption{Accuracy on Trad 5, Hadrian, Eclipse, ND-Twins, and CTTS.
ResNet 100 backbone for all matcher models.
``Trad 5'' is average accuracy across LFW, CALFW, CPLFW, CFP-FP and AgeDB-30.
Average and std dev from 10-fold cross-val are listed for Hadrian, Eclipse, ND-Twins and CTTS.
}
\label{tab:accuracies}
\end{table*}

Creation of CTTS began with lists of celebrity twins; e.g., \cite{SI_2012}.
Candidate MZ twins were dropped if a web search did not readily return at least 12 different images of each twin, with the face region being about 112x112 pixels or larger in each image, and with high-confidence identity labeling.
The minimum number of images per person and the minimum size for the face region favored twins with greater celebrity status and online presence.
High-confidence identity labeling favored twins who played for different sports teams, held different positions in government and / or had distinguishing facial marks visible in many images.

Of the 80 sets of MZ twins in CTTS, 43 have a skin mark that could be used to distinguish between them, and 7 have one twin that is left-handed and the other right-handed, indicating mirror twins. There are 36 male and 44 female sets of twins.  There is one set of twins born in the 1940s, 9 in the 1960s, 17 in the 1970s, 26 in the 1980s, 21 in the 1990s, and 6 in the early 2000s.

Web-scraping of publicly accessible data is and has long been legal in the US \cite{Whittaker_2022, QE_2023, Barrett_2025, Datashake_2026}. 
CTTS images are versions of face regions cropped from publicly-accessible images, with each face region normalized to standard orientation and 112x112 pixels in size, as appropriate for input to popular deep CNN matchers.
Images are organized into a balanced number of pairs, and pairs into folds, to create a test set appropriate to ten-fold cross-validation.
CTTS-80 is available to researchers who certify agreement to non-commercial use and that CTTS use is legal in their jurisdiction.
See {\tt https://github.com/mzang20/CTTS} for additional details and access to CTTS.

With 12 images selected for each individual, there are 132 same-person image pairs and 144 different-person pairs for each set of twins.
In the context of this type of face verification test set, same-person pairs are ``positive'' pairs and different-person pairs are ``negative'' pairs.
The number of positive and negative pairs for each set of twins is balanced by randomly selecting 132 of the 144 negative pairs.
Each fold contains 2x132= 264 image pairs for each of 8 sets of twins. Thus there are 264x8= 2,112 image pairs per fold and 21,120 image pairs total across the 10 folds.
This is over 3.5 times the 6,000 image pairs in other test sets LFW, CALFW, CPLFW, CFP-FP, AgeDB-30, Hadrian, Eclipse or ND-Twins.

There are two main approaches to reporting test accuracy for face verification.
One is to report FNMR at a threshold determined by a specific FMR on an impostor distribution.
This is the approach used in the NIST report \cite{hanaoka2022nist}, and associated with the IJB benchmarks \cite{Maze_2018}.
The other approach is to report the average accuracy on 10-fold cross-validation of a dataset of positive and negative image pairs.  
The threshold that gives highest accuracy classifying image pairs on 9 folds is used to classify the image pairs in the 10-th fold.  Each fold has accuracy computed in this way, and the average and standard deviation of the accuracies on the 10 folds is reported.
This is the approach used with LFW, CALFW, CPLFW, CFP-FP, AgeDB-30, Hadrian, Eclipse, ND-Twins, CTTS and various other face verification test sets.

The threshold used to classify image pairs in CTTS is selected based on classifying other twins image pairs, whereas the threshold used to classify twins image pairs in the NIST report is selected for a 1-in-10,000 FMR on a distribution of non-twins image pairs.
Due to the different approaches, it is possible that a face matcher could have a 0.99 FMR in the NIST results and also have 75\% accuracy on CTTS.

The CTTS folds are twin-set-disjoint, meaning that all image pairs of a given set of twins appear in the same fold.
Thus, when a given set of twins are scored for test accuracy, their image pairs are not used in selecting the threshold by which they are scored.
This should give an accuracy estimate that is less optimistically biased than if the image pairs for a given set of twins were distributed across all the folds.

Accuracy achieved on CTTS-80 by 4 matchers (ArcFace \cite{arcface}, AdaFace \cite{adaface}, UniFace \cite{uniface}, MagFace \cite{magface}), each trained on 3 different widely-used training sets (MS1MV2 \cite{arcface}, WebFace4M \cite{webface260m}, glint360k \cite{glint360k}), for a total of 12 matcher instances, is shown in Table \ref{tab:accuracies}.
For context, accuracy for the 12 matcher instances on the traditional five (``Trad 5'') face verification test sets LFW, CALFW, CPLFW, CFP-FP and AgeDB-30 is also shown, along with accuracy for newer test sets emphasizing facial hairstyles (``Hadrian''), lighting differences (``Eclipse'') and ND-Twins.
Accuracies in Table \ref{tab:accuracies} for the Trad 5 test sets follow the pattern expected from the literature, $\geq$ 97\% for all 12 matchers.
The test sets that are focused on challenging facial hair conditions (Hadrian) and illumination conditions (Eclipse) have lower accuracy than  Trad 5.
CTTS-80 and ND-Twins have still lower accuracies. 
It is important to note that the standard deviation is higher for the twins test sets, making it difficult to draw firm conclusions about accuracy differences between matchers, training sets, or ND-Twins versus CTTS-80.
Accuracy is more stable across matchers and training sets for CTTS than for ND-Twins, likely due to the larger number of image pairs in CTTS.

Available training sets for deep CNN face matchers, including MS1MV2, glint360k and WebFace 4M, are web-scraped.
CTTS is also web-scraped. 
MZ twins in CTTS who were established celebrities well before the training sets were created, such as Ronde and Tiki Barber or Aaron and Shawn Ashmore, likely also have images in the training sets.
Other CTTS twins, like Keegan and Kris Murray, and Chase and Sydney Brown, who were born in 2000 and have fairly recently become celebrity pro athletes, are almost certainly not in the circa 2019 MS1MV2 and likely also not in the other training sets.
It is possible in principle that a matcher could have higher accuracy for CTTS twins that are in the matcher's training set.
However, the overlapped twins between CTTS and a training set would be such a very small percent of the training set that it is unlikely there is any significant effect.

To examine the accuracy across different sets of twins in more detail, we selected the ArcFace matcher trained on MS1MV2 to use in results in the rest of this paper.
This matcher gives the highest accuracy on CTTS of the 12 matcher instances in Table~\ref{tab:accuracies}, though not by a statistically significant margin.
This matcher should give a representative idea of the accuracy that can be achieved by current matchers on CTTS.

We first examined how accuracy varies across the sets of twins.
The standard deviation listed in Table~\ref{tab:accuracies} is for the average accuracy across the 10 folds.
The range of accuracy variation across the 80 sets of twins is surprisingly large: 7 sets of twins had accuracy in the range (50\%, 60\%), 19 in [60\%, 70\%), 26 in [70\%, 80\%), 11 in [80\%, 90\%) and 17 in [90\%, 100\%).

Accuracy for a set of twins results from two factors.
One, different twins can have different levels of separation in the distributions of the distances for their positive and negative image pairs.
The separation in their distributions determines the max potential accuracy for a set of twins.
Two, the decision threshold selected based on the other 9 folds may or may not generalize well to a given set of twins in the 10-th fold.
Twins with high accuracy have good separation of their distributions and a well-placed threshold.
Twins with low accuracy have poor separation of their distributions, a poorly-placed threshold, or both.

\begin{figure*}
\centering
\includegraphics[width=0.15\textwidth]{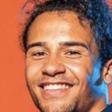}
\includegraphics[width=0.15\textwidth]{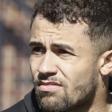}
\includegraphics[width=0.15\textwidth]{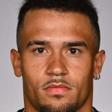}
\includegraphics[width=0.15\textwidth]{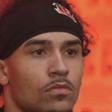}
\includegraphics[width=0.15\textwidth]{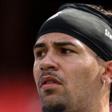}
\includegraphics[width=0.15\textwidth]{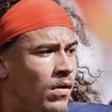}
\\
\includegraphics[width=0.15\textwidth]{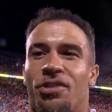}
\includegraphics[width=0.15\textwidth]{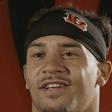}
\includegraphics[width=0.15\textwidth]{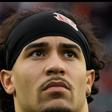}
\includegraphics[width=0.15\textwidth]{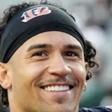}
\includegraphics[width=0.15\textwidth]{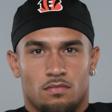}
\includegraphics[width=0.15\textwidth]{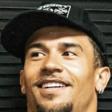}
\\
(a) Twelve 112x112 normalized face crops - Chase Brown
\\[6pt]
\includegraphics[width=0.15\textwidth]{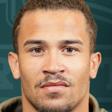}
\includegraphics[width=0.15\textwidth]{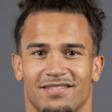}
\includegraphics[width=0.15\textwidth]{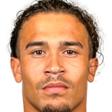}
\includegraphics[width=0.15\textwidth]{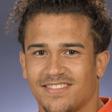}
\includegraphics[width=0.15\textwidth]{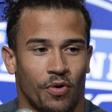}
\includegraphics[width=0.15\textwidth]{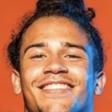}
\\
\includegraphics[width=0.15\textwidth]{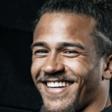}
\includegraphics[width=0.15\textwidth]{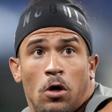}
\includegraphics[width=0.15\textwidth]{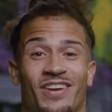}
\includegraphics[width=0.15\textwidth]{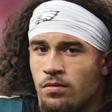}
\includegraphics[width=0.15\textwidth]{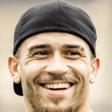}
\includegraphics[width=0.15\textwidth]{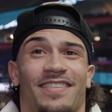}
\\
(b) Twelve 112x112 normalized face crops - Sydney Brown
\\[6pt]
\includegraphics[width=0.85
\textwidth]{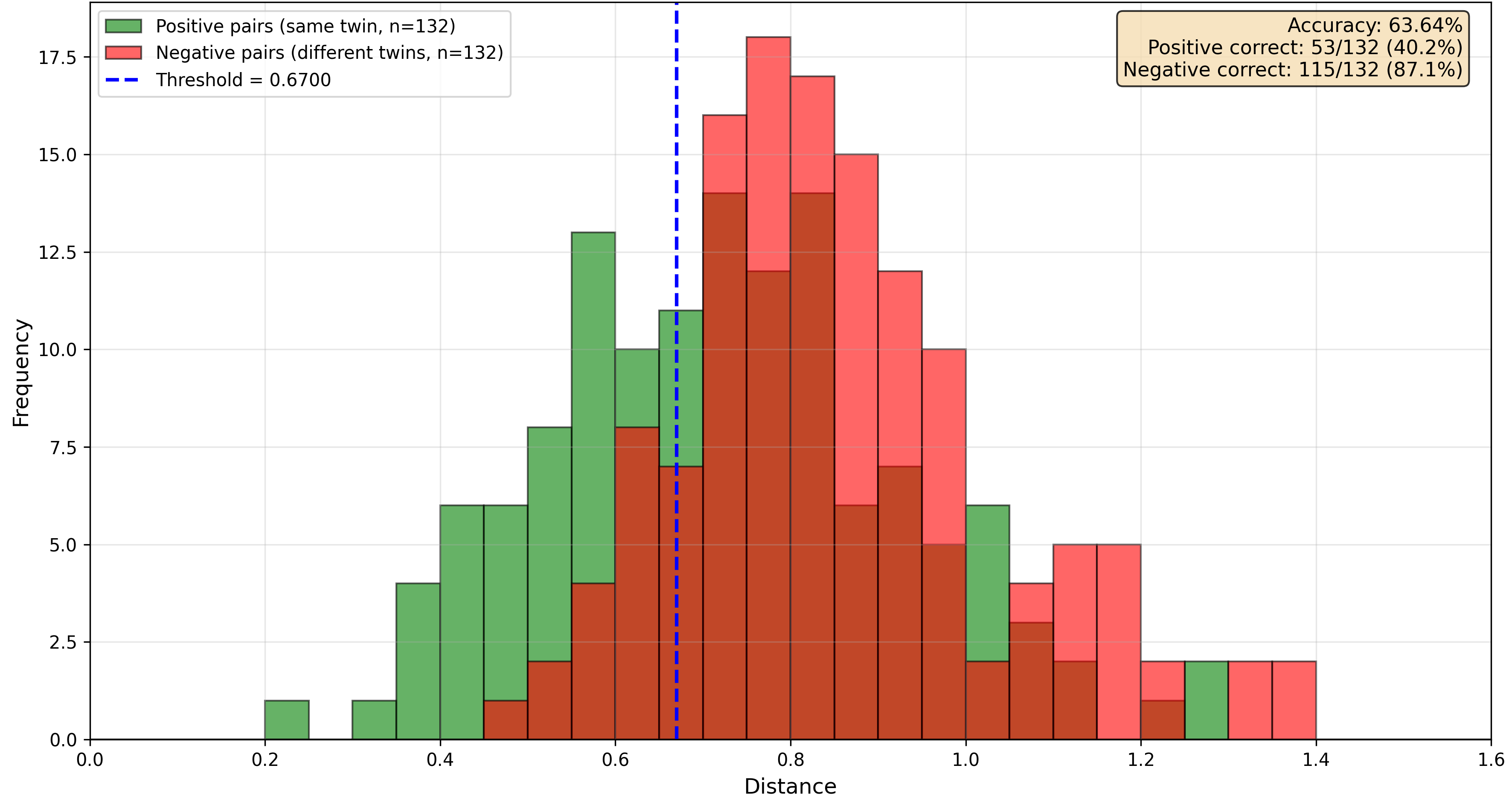}
\\
(c) Distance Distributions of Same-Person (``positive'') and Different-Person (``negative'') Image Pairs
\caption{Example of Not-Well-Separated Distance Score Distributions. 
Euclidean distance of embeddings from ArcFace ResNet-100 backbone trained on MS1MV2.
Chase is left-handed and Sydney is right-handed, indicating mirror twins.}
\label{fig:Brown_twins}
\end{figure*}

\begin{figure*}
\centering
\includegraphics[width=0.15\textwidth]{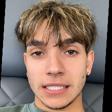}
\includegraphics[width=0.15\textwidth]{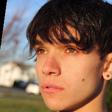}
\includegraphics[width=0.15\textwidth]{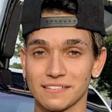}
\includegraphics[width=0.15\textwidth]{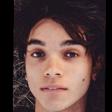}
\includegraphics[width=0.15\textwidth]{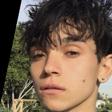}
\includegraphics[width=0.15\textwidth]{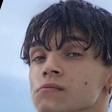}
\\
\includegraphics[width=0.15\textwidth]{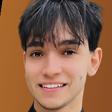}
\includegraphics[width=0.15\textwidth]{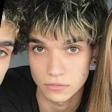}
\includegraphics[width=0.15\textwidth]{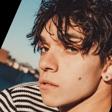}
\includegraphics[width=0.15\textwidth]{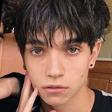}
\includegraphics[width=0.15\textwidth]{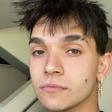}
\includegraphics[width=0.15\textwidth]{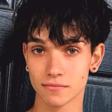}
\\
(a) Twelve 112x112 normalized face crops - Lucas Dobre
\\[6pt]
\includegraphics[width=0.15\textwidth]{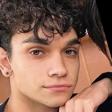}
\includegraphics[width=0.15\textwidth]{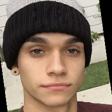}
\includegraphics[width=0.15\textwidth]{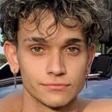}
\includegraphics[width=0.15\textwidth]{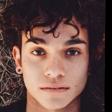}
\includegraphics[width=0.15\textwidth]{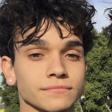}
\includegraphics[width=0.15\textwidth]{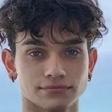}
\\
\includegraphics[width=0.15\textwidth]{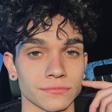}
\includegraphics[width=0.15\textwidth]{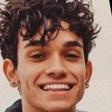}
\includegraphics[width=0.15\textwidth]{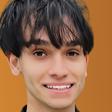}
\includegraphics[width=0.15\textwidth]{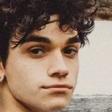}
\includegraphics[width=0.15\textwidth]{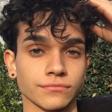}
\includegraphics[width=0.15\textwidth]{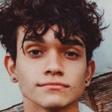}
\\
(b) Twelve 112x112 normalized face crops - Marcus Dobre
\\[6pt]
\includegraphics[width=0.85
\textwidth]{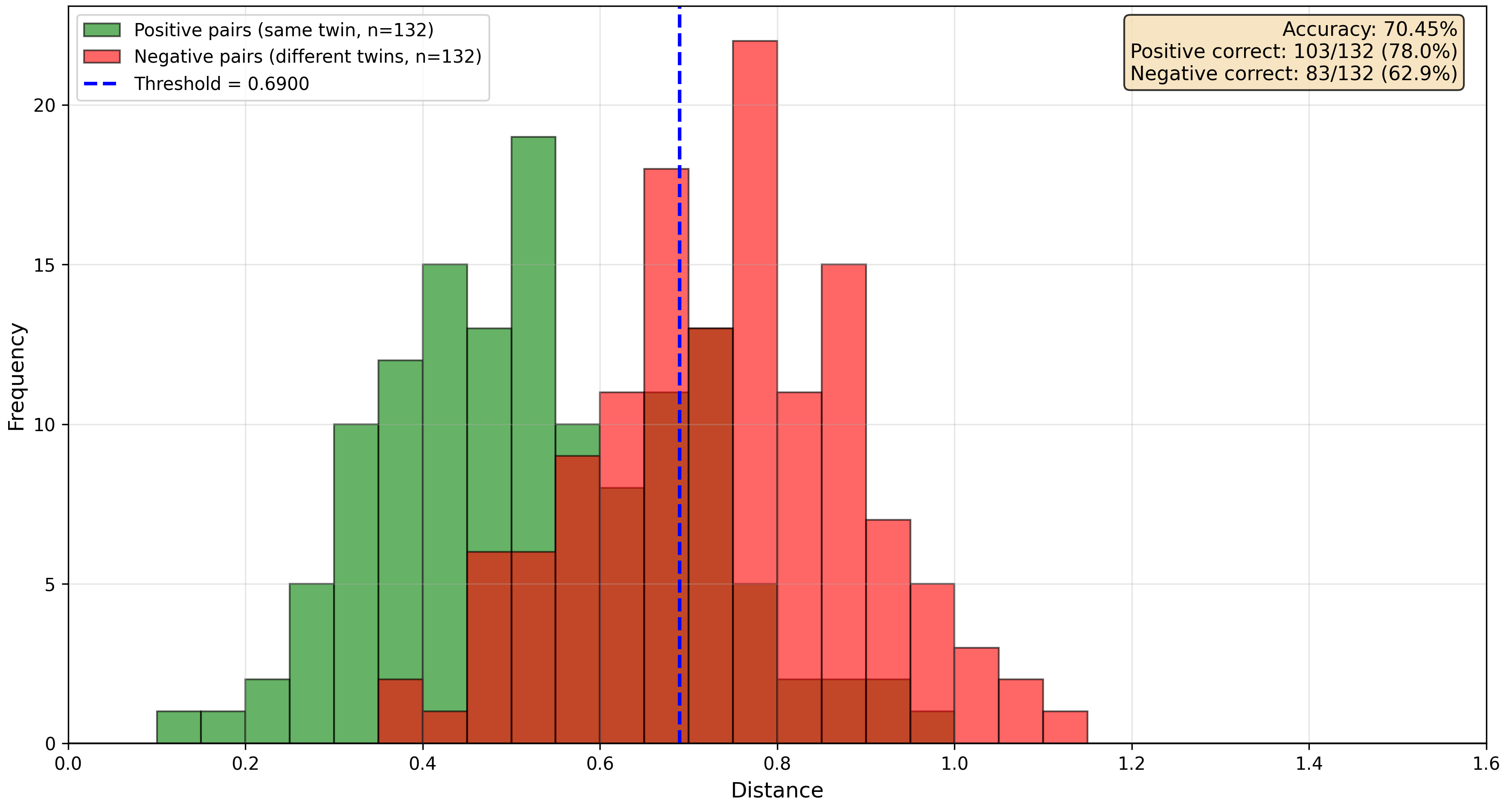}
\\
(c) Distance Distributions of Same-Person (``positive'') and Different-Person (``negative'') Image Pairs
\caption{Example of Not-Well-Separated Distance Score Distributions for MZ Twins. 
Euclidean distance of embeddings from ArcFace ResNet-100 backbone trained on MS1MV2.  
Threshold selected from other 9 folds in 10-fold cross-val is higher than ideal for these distributions.  Note that Lucas Dobre has a skin mark on his left cheek, visible in these images, that  distinguishes between the twins.}
\label{fig:Dobre_twins}
\end{figure*}

\begin{figure*}
\centering
\includegraphics[width=0.15\textwidth]{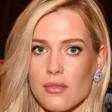}
\includegraphics[width=0.15\textwidth]{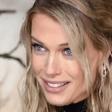}
\includegraphics[width=0.15\textwidth]{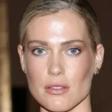}
\includegraphics[width=0.15\textwidth]{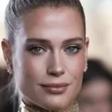}
\includegraphics[width=0.15\textwidth]{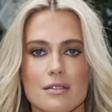}
\includegraphics[width=0.15\textwidth]{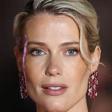}
\\
\includegraphics[width=0.15\textwidth]{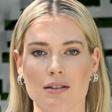}
\includegraphics[width=0.15\textwidth]{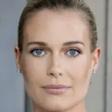}
\includegraphics[width=0.15\textwidth]{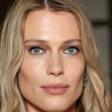}
\includegraphics[width=0.15\textwidth]{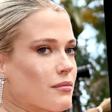}
\includegraphics[width=0.15\textwidth]{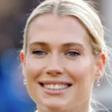}
\includegraphics[width=0.15\textwidth]{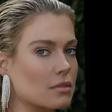}
\\
(a) Twelve 112x112 normalized face crops - Amelia Spencer
\\[6pt]
\includegraphics[width=0.15\textwidth]{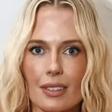}
\includegraphics[width=0.15\textwidth]{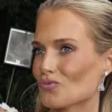}
\includegraphics[width=0.15\textwidth]{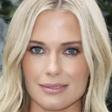}
\includegraphics[width=0.15\textwidth]{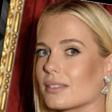}
\includegraphics[width=0.15\textwidth]{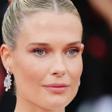}
\includegraphics[width=0.15\textwidth]{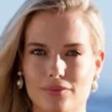}
\\
\includegraphics[width=0.15\textwidth]{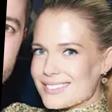}
\includegraphics[width=0.15\textwidth]{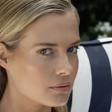}
\includegraphics[width=0.15\textwidth]{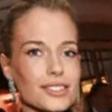}
\includegraphics[width=0.15\textwidth]{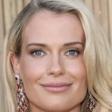}
\includegraphics[width=0.15\textwidth]{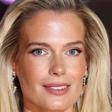}
\includegraphics[width=0.15\textwidth]{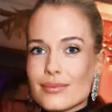}
\\
(b) Twelve 112x112 normalized face crops - Eliza Spencer
\\[6pt]
\includegraphics[width=0.85
\textwidth]{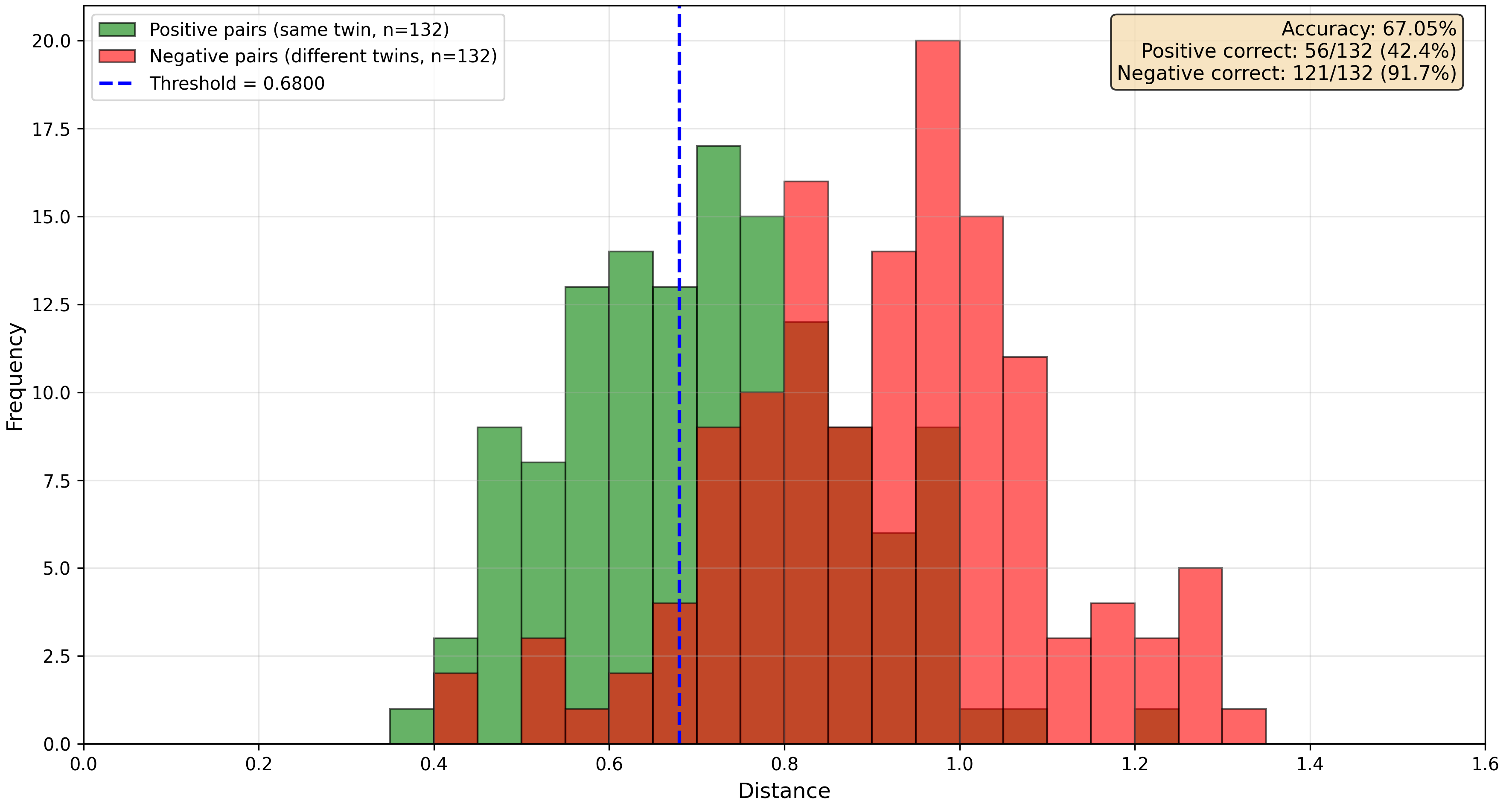}
\\
(c) Distance Distributions of Same-Person (``positive'') and Different-Person (``negative'') Image Pairs
\caption{Example of Not-Well-Separated Distance Score Distributions. 
Euclidean distance of embeddings from ArcFace ResNet-100 backbone trained on MS1MV2.
Threshold selected from other 9 folds in 10-fold cross-val is lower than ideal for these distributions.   
}
\label{fig:Spencer_twins}
\end{figure*}

Examples of well-separated distance distributions appear in Figures \ref{fig:Barber_twins}, \ref{fig:Goss_twins} and \ref{fig:Mowry_twins}.
Note that the faces in these examples generally have approximately frontal pose, little or no occlusion by hair, no hats and eyeglasses in only one image.
Examples of not-well-separated distance distributions appear in Figures \ref{fig:Brown_twins}, \ref{fig:Dobre_twins} and \ref{fig:Spencer_twins}.
Multiple images of each twin in Figure \ref{fig:Brown_twins} contain occlusion by caps, and there are examples of off-frontal pose and atypical facial expressions.
Images in Figure \ref{fig:Dobre_twins} show substantial occlusion of the forehead by hairstyle and caps, and substantial pose variation.
Images in Figure \ref{fig:Spencer_twins} show substantial variation in pose and in hairstyle.
This contrast in the image variation between twins with well-separated and not-well-separated distance distributions suggests that it may be possible to  design image acquisition to get well-separated distance distributions.

\section{Do Current Matchers Encode Skin Marks?}
\label{sec:skin_marks}

\begin{figure*}
\centering
\includegraphics[width=0.5\textwidth]{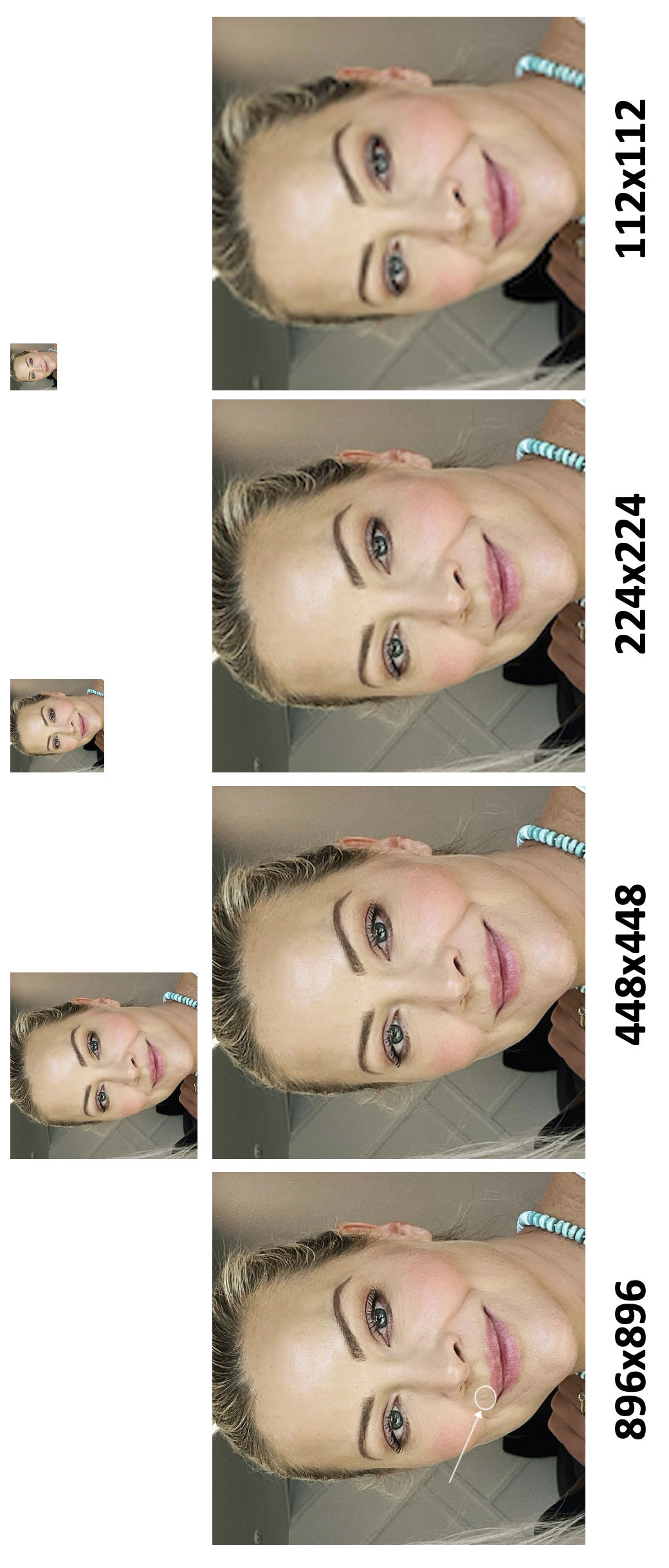}
\caption{Skin Mark Visibility And Image Resolution.
Upper row of images shown at size on the page to illustrate differences in resolution.
Lower row of images shown at same size on the page to illustrate changing visibility of skin mark.
}
\label{fig:Cynthia_resolutions}
\end{figure*}

As described earlier, previous research has emphasized skin marks as a means to distinguish identical twins \cite{biswas2011study, Srinivas2012, Le_2012, Le_2015, Mousavi_2021, Sundaresan_2021, hanaoka2022nist, Brahmbhatt2026}.
Over half the twins in CTTS have skin marks that can be used to distinguish them.
Of course, skin marks may not be visible in some images, due to pose, cosmetics, image resolution or blur.
Figure \ref{fig:Cynthia_resolutions} shows how skin marks that are easily visualized in higher resolution images can become ambiguous at the 112x112 resolution commonly used by face matching models.
However, distinguishing skin marks are visible in the 112x112 pixel images of the twins in Figures \ref{fig:Mowry_twins}, \ref{fig:Dobre_twins}, \ref{fig:Hassan_twins} and \ref{fig:Kaczynski_twins}.
This raises the question - Do current deep CNN matchers make use of skin marks?

\begin{figure*}
\centering
\includegraphics[width=0.15\textwidth]{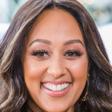}
\includegraphics[width=0.15\textwidth]{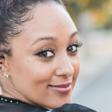}
\includegraphics[width=0.15\textwidth]{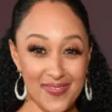}
\includegraphics[width=0.15\textwidth]{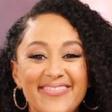}
\includegraphics[width=0.15\textwidth]{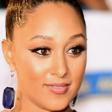}
\includegraphics[width=0.15\textwidth]{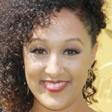}
\\
\includegraphics[width=0.15\textwidth]{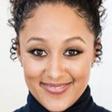}
\includegraphics[width=0.15\textwidth]{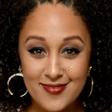}
\includegraphics[width=0.15\textwidth]{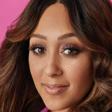}
\includegraphics[width=0.15\textwidth]{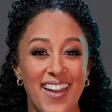}
\includegraphics[width=0.15\textwidth]{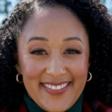}
\includegraphics[width=0.15\textwidth]{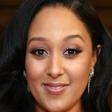}
\\
(a) Skin-mark-erased Versions of Images in Figure \ref{fig:Mowry_twins}(a)
\\[6pt]

\includegraphics[width=0.7
\textwidth]{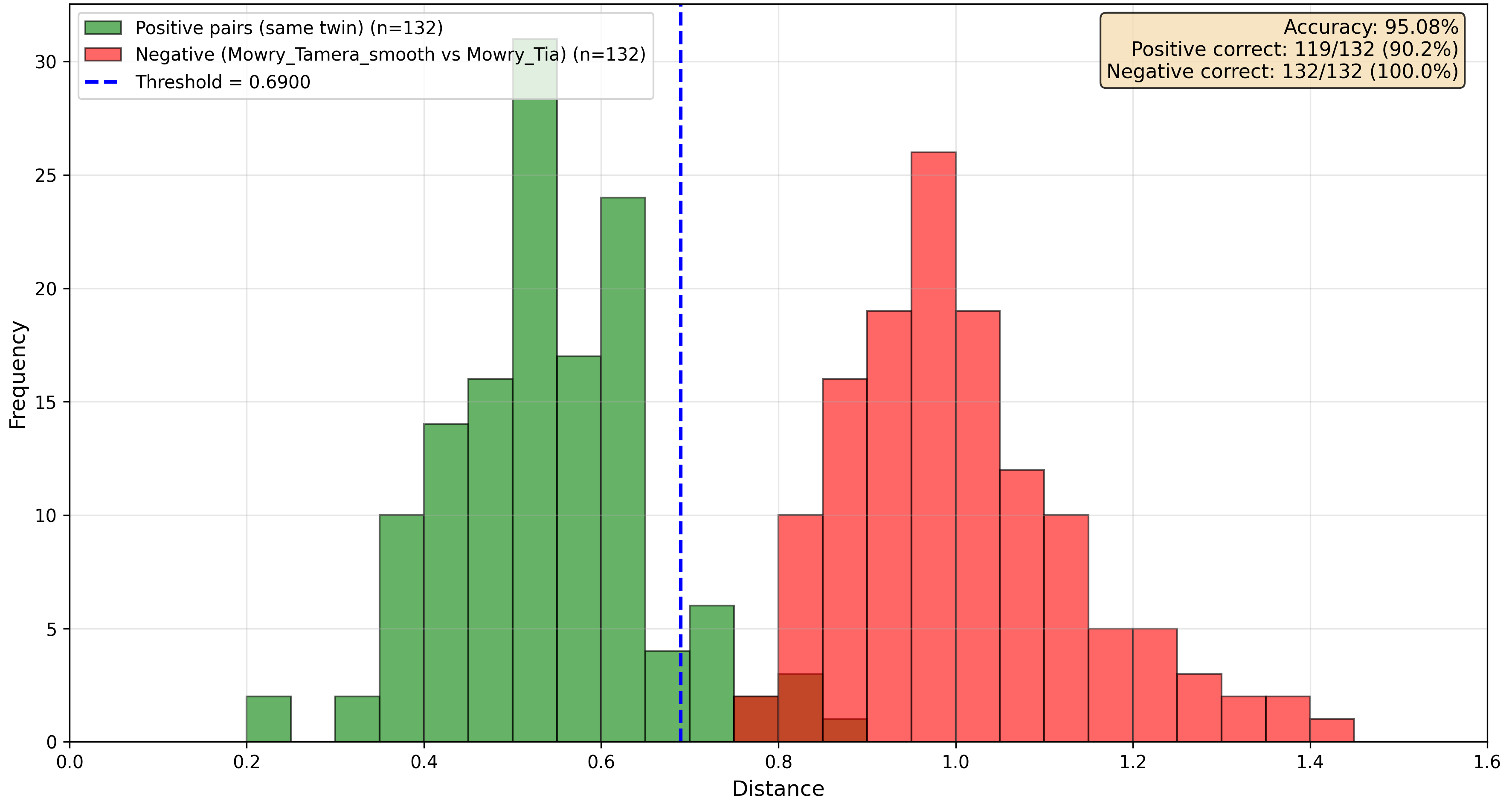}
\\
(b) Distance Distributions Using Skin-mark-erased Tamera Images; compare to Figure \ref{fig:Mowry_twins}(c)
\\[6pt]

\includegraphics[width=0.7
\textwidth]{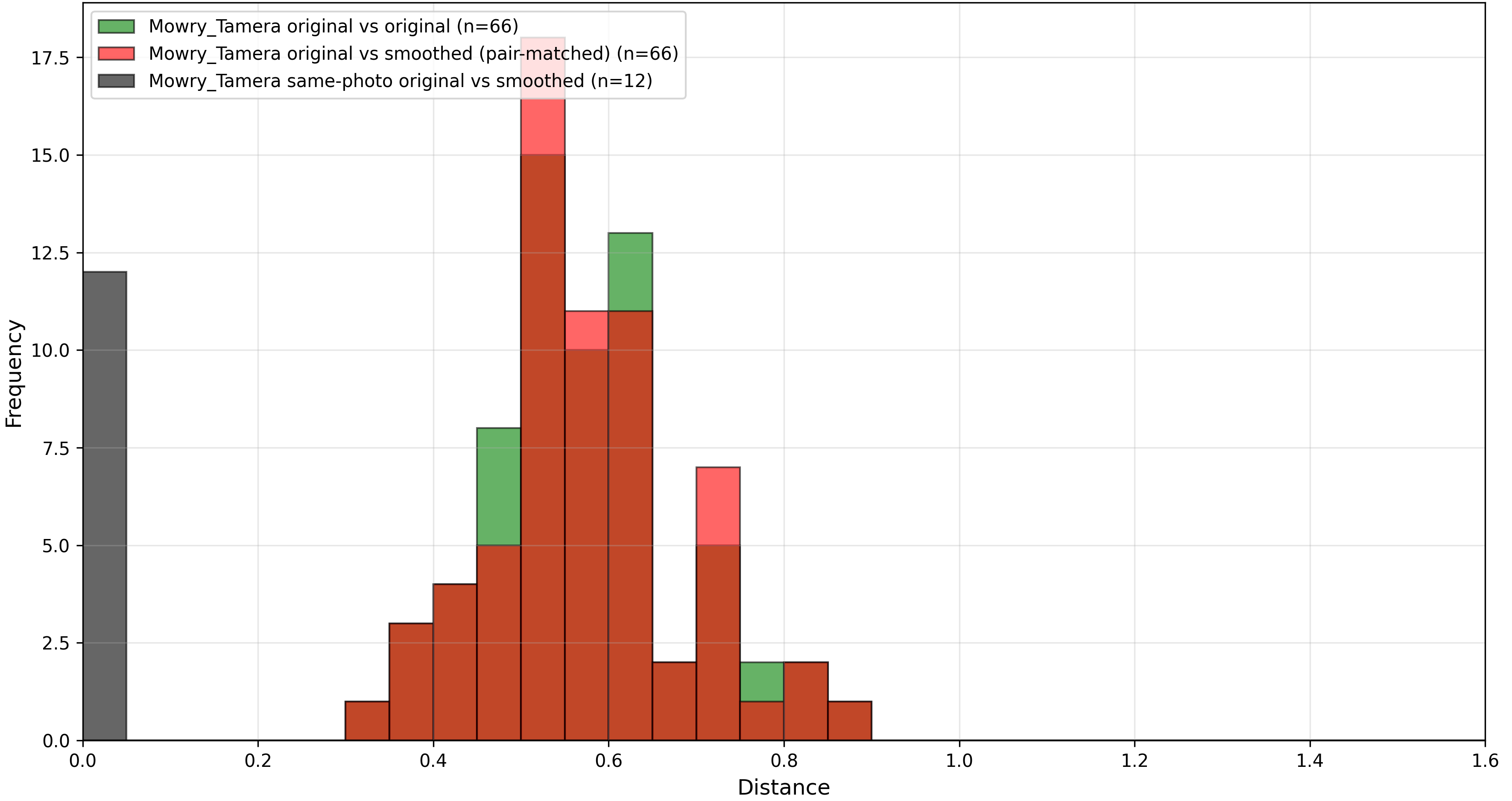}
\\
(c) Distance Distributions from Matching Original and Skin-mark-erased Images 
\caption{Exploring the Effect of Skin-Mark-Erased Images. Using skin-mark-erased images produces essentially the same result as images with skin mark present.
Matching skin-mark-erased to skin-mark-present images results in distances clustered near zero, indicating that the skin mark has essentially no impact on the embedding computed by the model.
}
\label{fig:Tamera_smoothed}
\end{figure*}

\begin{figure*}
\centering
\includegraphics[width=0.15\textwidth]{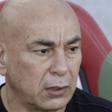}
\includegraphics[width=0.15\textwidth]{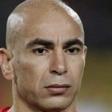}
\includegraphics[width=0.15\textwidth]{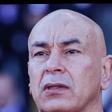}
\includegraphics[width=0.15\textwidth]{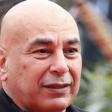}
\includegraphics[width=0.15\textwidth]{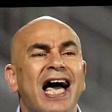}
\includegraphics[width=0.15\textwidth]{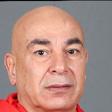}
\\
\includegraphics[width=0.15\textwidth]{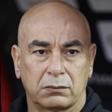}
\includegraphics[width=0.15\textwidth]{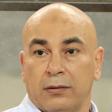}
\includegraphics[width=0.15\textwidth]{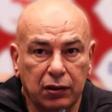}
\includegraphics[width=0.15\textwidth]{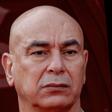}
\includegraphics[width=0.15\textwidth]{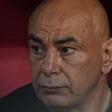}
\includegraphics[width=0.15\textwidth]{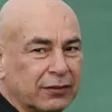}
\\
(a) Twelve 112x112 normalized face crops - Hossam Hassan
\\[6pt]
\includegraphics[width=0.15\textwidth]{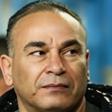}
\includegraphics[width=0.15\textwidth]{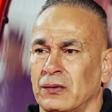}
\includegraphics[width=0.15\textwidth]{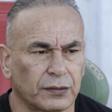}
\includegraphics[width=0.15\textwidth]{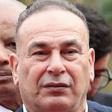}
\includegraphics[width=0.15\textwidth]{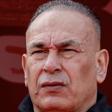}
\includegraphics[width=0.15\textwidth]{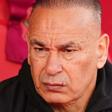}
\\
\includegraphics[width=0.15\textwidth]{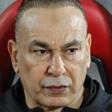}
\includegraphics[width=0.15\textwidth]{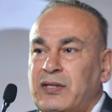}
\includegraphics[width=0.15\textwidth]{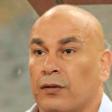}
\includegraphics[width=0.15\textwidth]{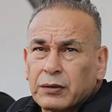}
\includegraphics[width=0.15\textwidth]{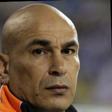}
\includegraphics[width=0.15\textwidth]{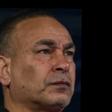}
\\
(b) Twelve 112x112 normalized face crops - Ibrahim Hassan
\\[6pt]
\includegraphics[width=0.85
\textwidth]{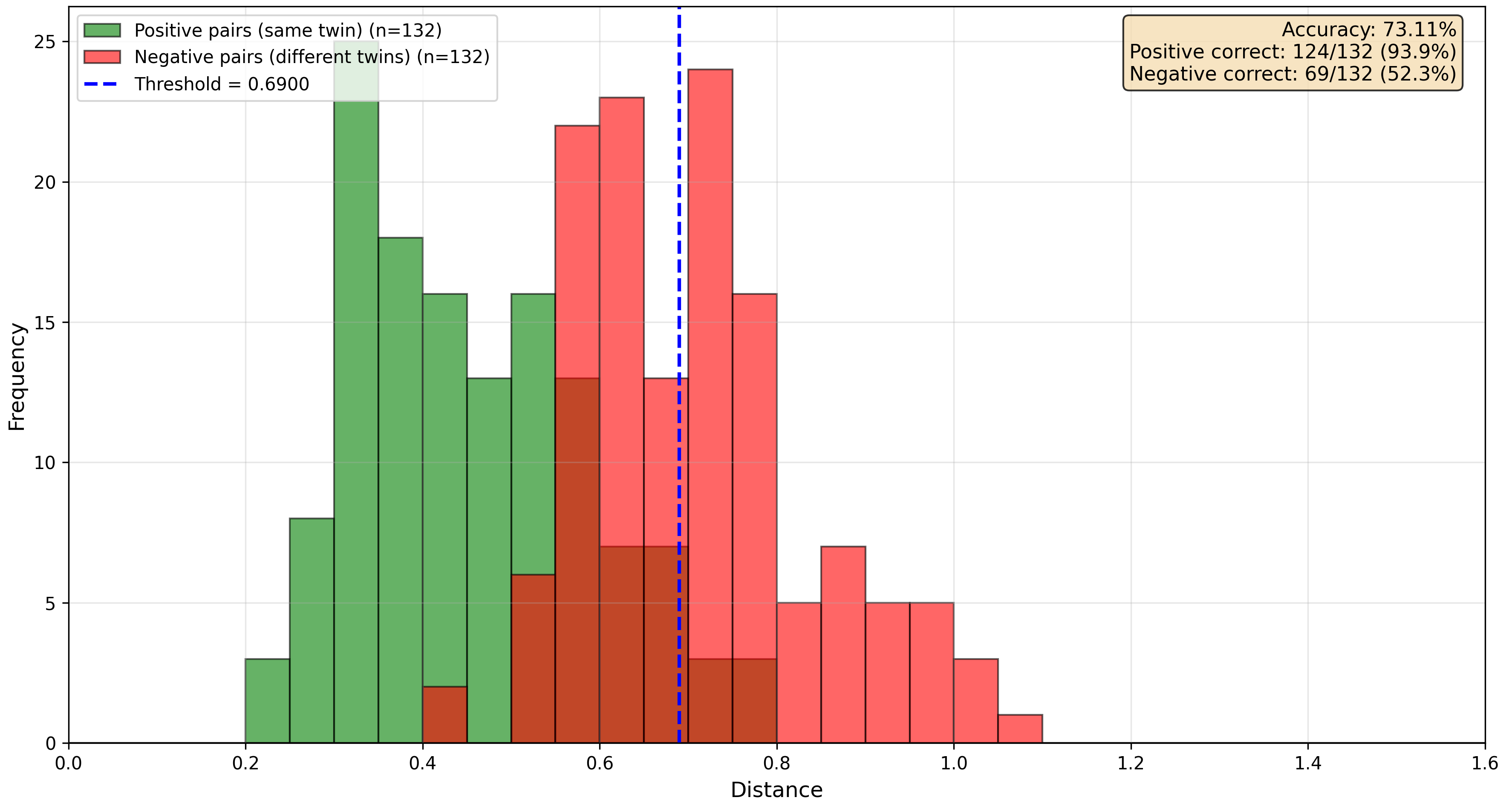}
\\
(c) Distance Distributions of Same-Person (``positive'') and Different-Person (``negative'') Image Pairs
\caption{Example of Distance Score Distributions. 
Euclidean distance of embeddings from ArcFace ResNet-100 backbone trained on MS1MV2.
Ibrahim Hassan has a skin mark visible in these images outside and below the right corner of his lips that distinguishes the twins.
}
\label{fig:Hassan_twins}
\end{figure*}

For an experimental answer to this question, we created skin-mark-erased versions of the Tamera Mowry images, editing only the area of the skin mark.
We then matched the skin-mark-erased versions of the Tamera images against the same Tia images as in Figure \ref{fig:Mowry_twins}.
If the skin-mark-visible images play a role in creating the well-separated distributions in Figure \ref{fig:Mowry_twins}, then the distributions with the skin-mark-erased versions of Tamera should be less well separated.
However, the distributions do not change significantly using the skin-mark-erased images; see Figure \ref{fig:Tamera_smoothed}.
This suggests that the visible skin mark has little or no effect on the embedding computed from the image. 

To investigate this more directly, we matched the original Tamera images against the skin-mark-erased Tamera images.
In this result, the positive pairs are images that either both have a skin mark or both do not, and the negative pairs have one image with a visible skin mark and one without.
The distance distributions for positive and negative pairs are nearly totally overlapped.
The exception is the 12 negative pairs that are an image with and without the visible skin mark.
The distances for these pairs are clustered near zero distance; see Figure \ref{fig:Tamera_smoothed}(c).
This is direct evidence that the  presence of the visible skin mark has almost no effect on the embedding computed from the image.

\begin{figure*}
\centering
\includegraphics[width=0.15\textwidth]{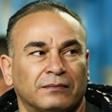}
\includegraphics[width=0.15\textwidth]{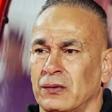}
\includegraphics[width=0.15\textwidth]{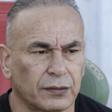}
\includegraphics[width=0.15\textwidth]{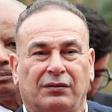}
\includegraphics[width=0.15\textwidth]{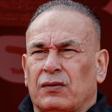}
\includegraphics[width=0.15\textwidth]{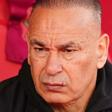}
\\
\includegraphics[width=0.15\textwidth]{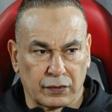}
\includegraphics[width=0.15\textwidth]{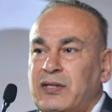}
\includegraphics[width=0.15\textwidth]{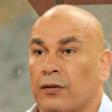}
\includegraphics[width=0.15\textwidth]{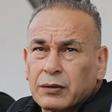}
\includegraphics[width=0.15\textwidth]{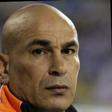}
\includegraphics[width=0.15\textwidth]{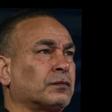}
\\
(a) Skin-mark-erased Versions of Images in Figure \ref{fig:Hassan_twins}(b)
\\[6pt]

\includegraphics[width=0.7
\textwidth]{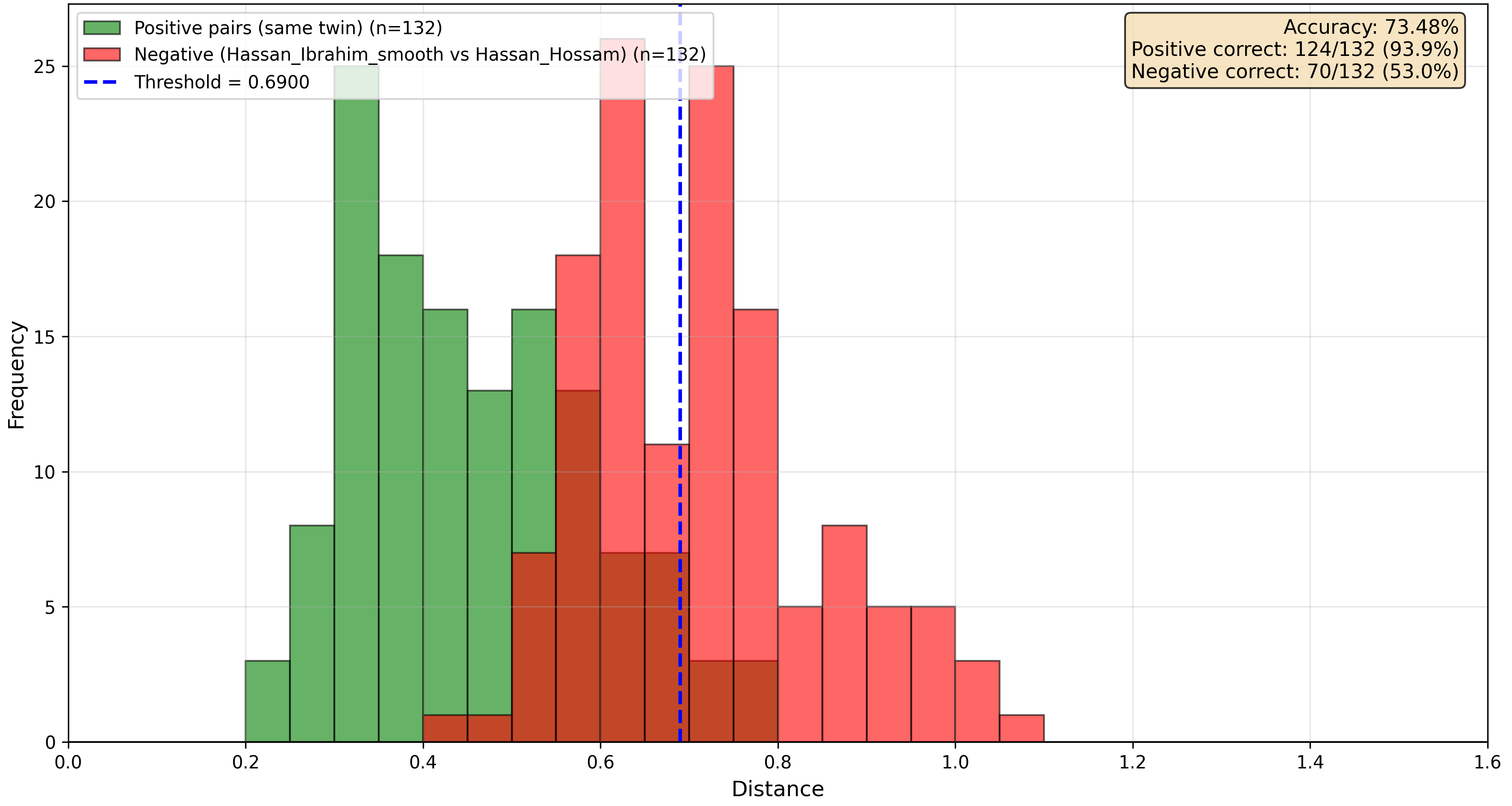}
\\
(b) Distance Distributions Using Skin-mark-erased Ibrahim Hassan Images; compare to Figure \ref{fig:Hassan_twins}(c)
\\[6pt]

\includegraphics[width=0.7
\textwidth]{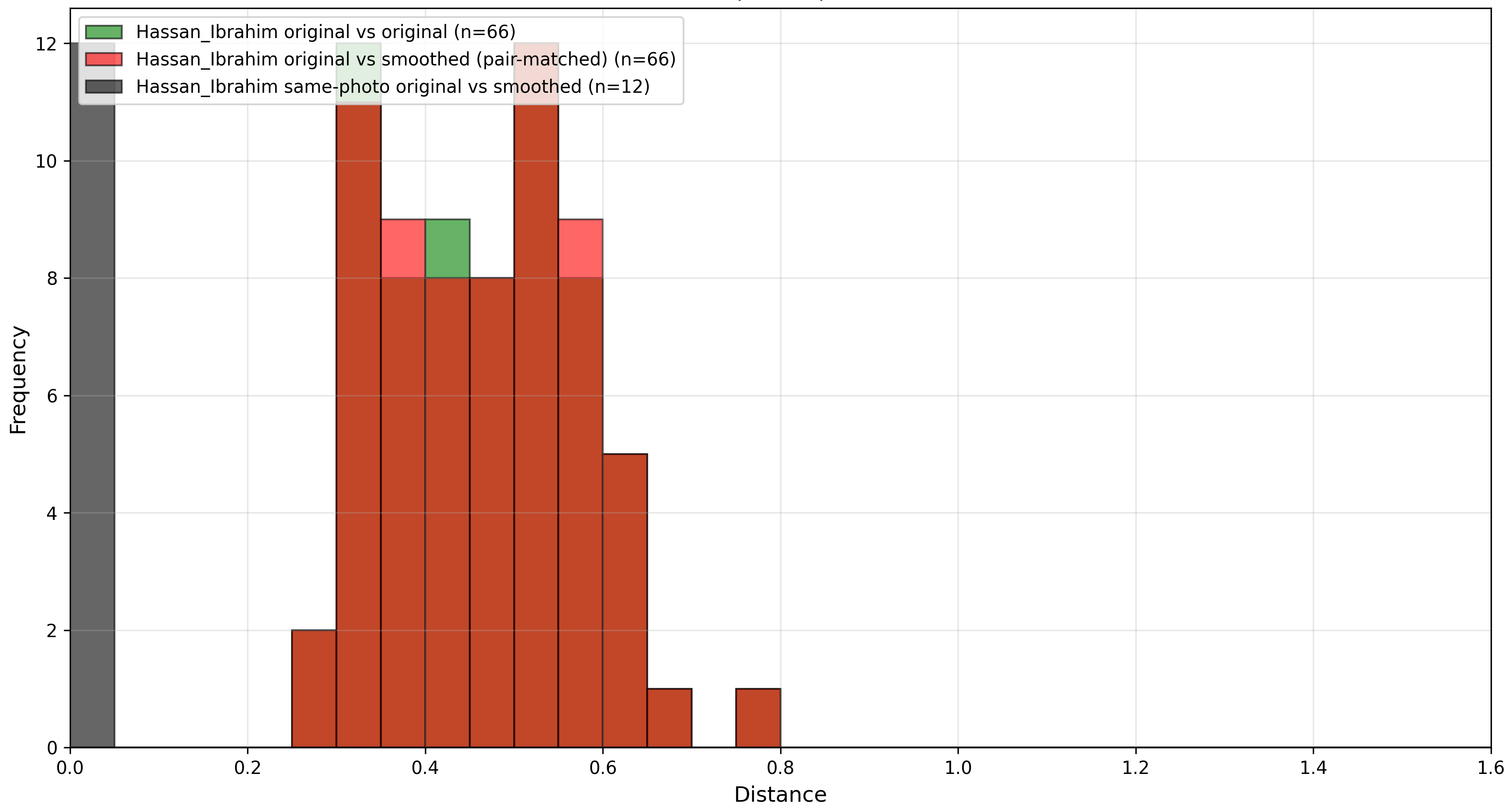}
\\
(c) Distance Distributions from Matching Original and Skin-mark-erased Images 
\caption{Exploring the Effect of Skin-Mark-Erased Images. Using skin-mark-erased images produces essentially the same result as images with skin mark present.
Matching skin-mark-erased to skin-mark-present images results in distances clustered near zero, indicating that the skin mark has essentially no impact on the embedding computed by the model.
}
\label{fig:Hassan_smoothed}
\end{figure*}

\begin{figure*}
\centering
\includegraphics[width=0.15\textwidth]{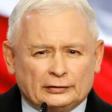}
\includegraphics[width=0.15\textwidth]{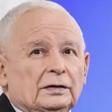}
\includegraphics[width=0.15\textwidth]{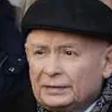}
\includegraphics[width=0.15\textwidth]{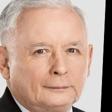}
\includegraphics[width=0.15\textwidth]{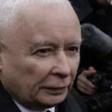}
\includegraphics[width=0.15\textwidth]{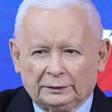}
\\
\includegraphics[width=0.15\textwidth]{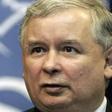}
\includegraphics[width=0.15\textwidth]{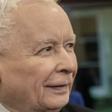}
\includegraphics[width=0.15\textwidth]{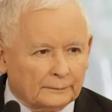}
\includegraphics[width=0.15\textwidth]{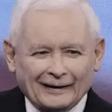}
\includegraphics[width=0.15\textwidth]{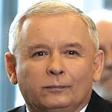}
\includegraphics[width=0.15\textwidth]{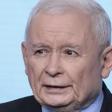}
\\
(a) Twelve 112x112 normalized face crops - Jaroslaw Kaczynski
\\[6pt]
\includegraphics[width=0.15\textwidth]{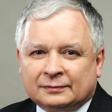}
\includegraphics[width=0.15\textwidth]{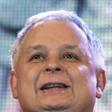}
\includegraphics[width=0.15\textwidth]{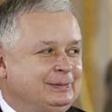}
\includegraphics[width=0.15\textwidth]{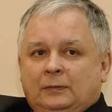}
\includegraphics[width=0.15\textwidth]{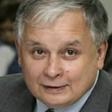}
\includegraphics[width=0.15\textwidth]{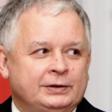}
\\
\includegraphics[width=0.15\textwidth]{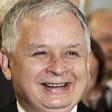}
\includegraphics[width=0.15\textwidth]{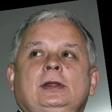}
\includegraphics[width=0.15\textwidth]{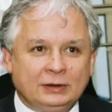}
\includegraphics[width=0.15\textwidth]{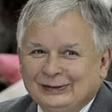}
\includegraphics[width=0.15\textwidth]{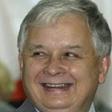}
\includegraphics[width=0.15\textwidth]{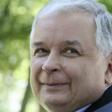}
\\
(b) Twelve 112x112 normalized face crops - Lech Kaczynski
\\[6pt]
\includegraphics[width=0.85
\textwidth]{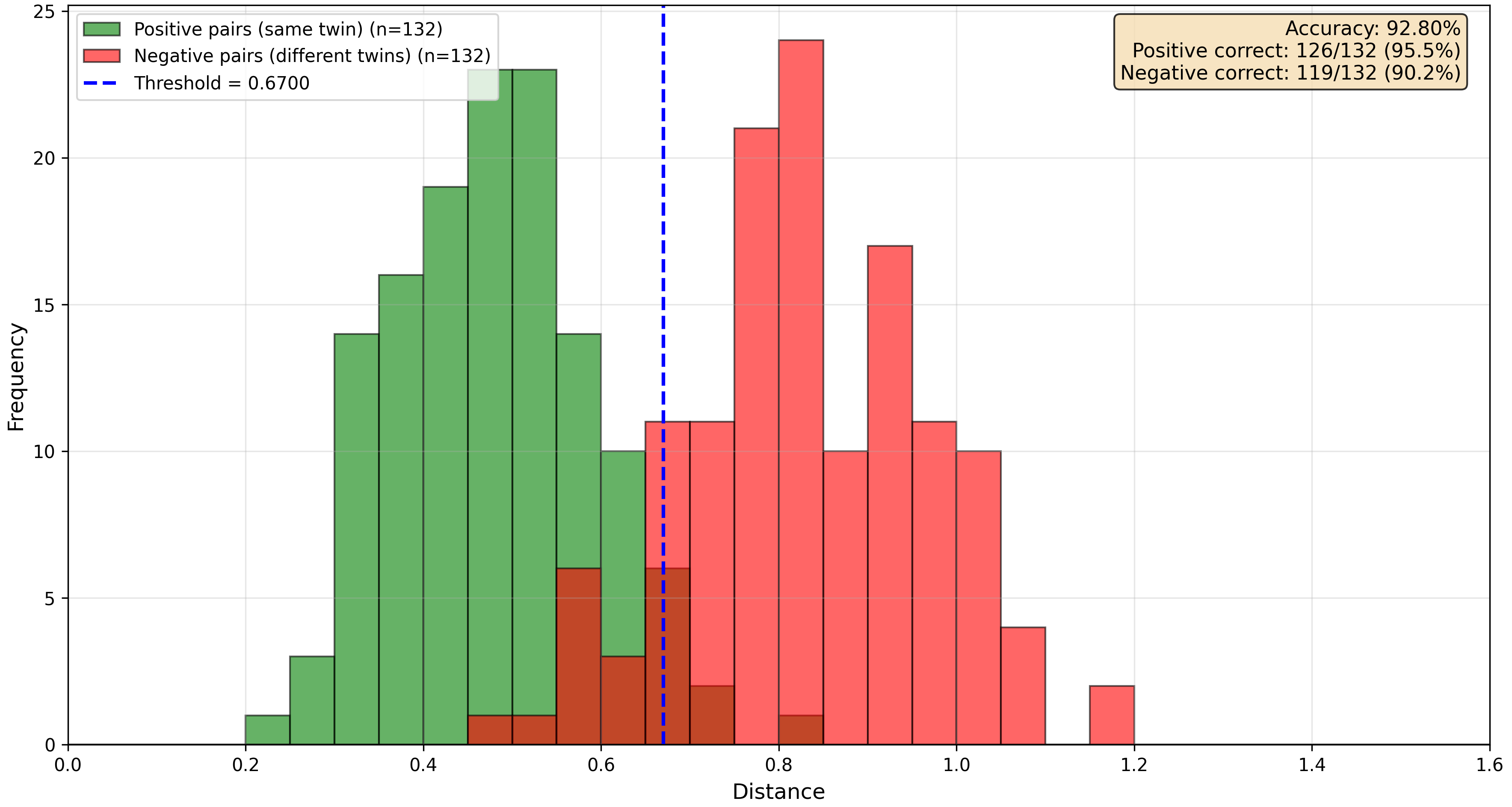}
\\
(c) Distance Distributions of Same-Person (``positive'') and Different-Person (``negative'') Image Pairs
\caption{Example of Distance Score Distributions. 
Euclidean distance of embeddings from ArcFace ResNet-100 backbone trained on MS1MV2.
Lech Kaczynski has a skin mark visible in these images below his left eye that distinguishes the twins.
}
\label{fig:Kaczynski_twins}
\end{figure*}

\begin{figure*}
\centering
\includegraphics[width=0.15\textwidth]{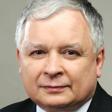}
\includegraphics[width=0.15\textwidth]{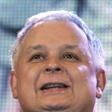}
\includegraphics[width=0.15\textwidth]{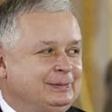}
\includegraphics[width=0.15\textwidth]{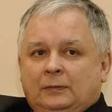}
\includegraphics[width=0.15\textwidth]{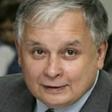}
\includegraphics[width=0.15\textwidth]{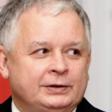}
\\
\includegraphics[width=0.15\textwidth]{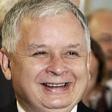}
\includegraphics[width=0.15\textwidth]{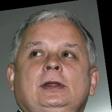}
\includegraphics[width=0.15\textwidth]{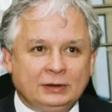}
\includegraphics[width=0.15\textwidth]{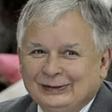}
\includegraphics[width=0.15\textwidth]{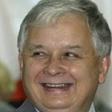}
\includegraphics[width=0.15\textwidth]{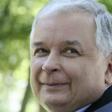}
\\
(a) Skin-mark-erased Versions of Images in Figure \ref{fig:Kaczynski_twins}(b)
\\[6pt]

\includegraphics[width=0.7
\textwidth]{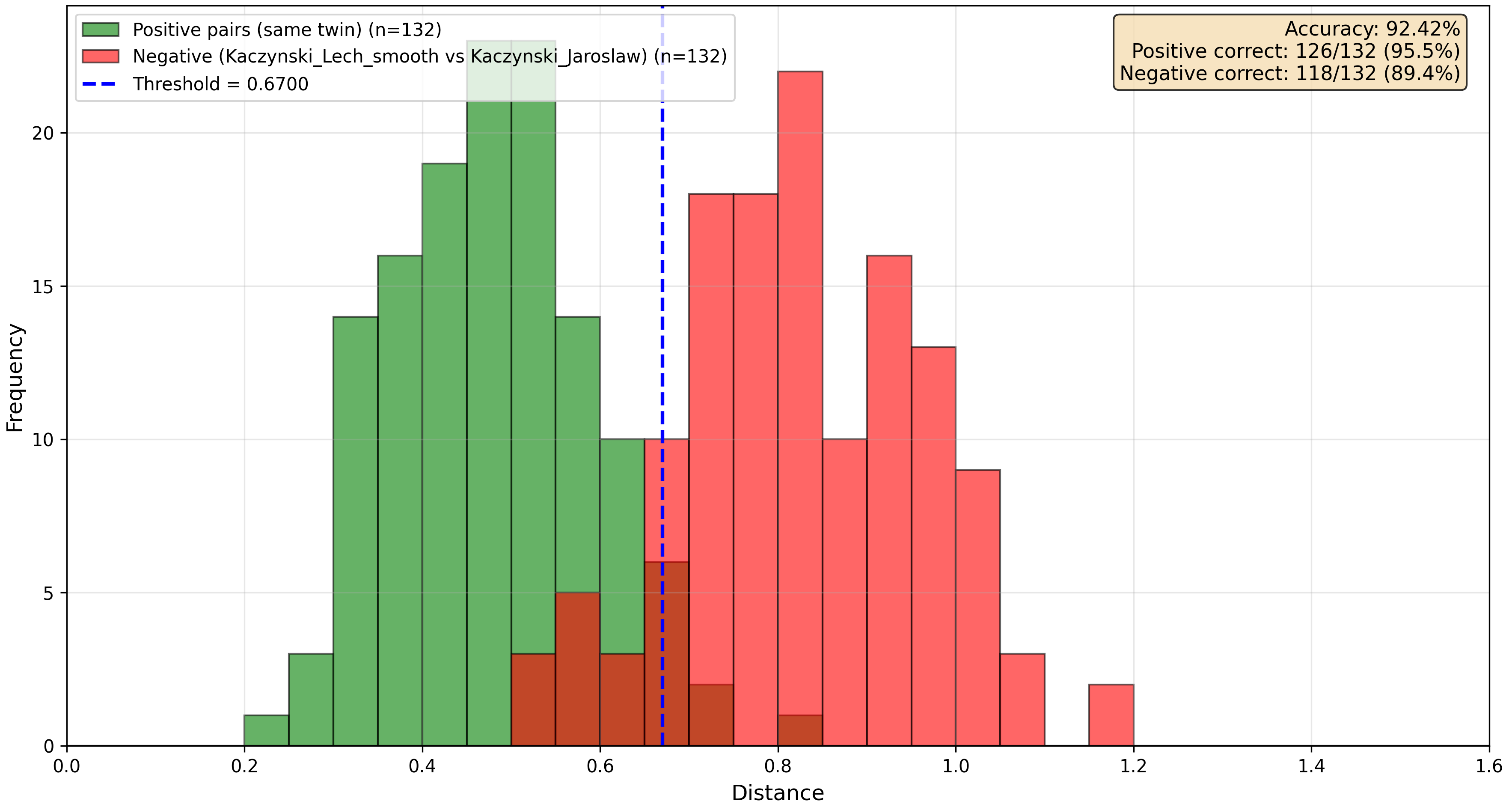}
\\
(b) Distance Distributions Using Skin-mark-erased Lech Kaczynski Images; compare to Figure \ref{fig:Kaczynski_twins}(c)
\\[6pt]

\includegraphics[width=0.7
\textwidth]{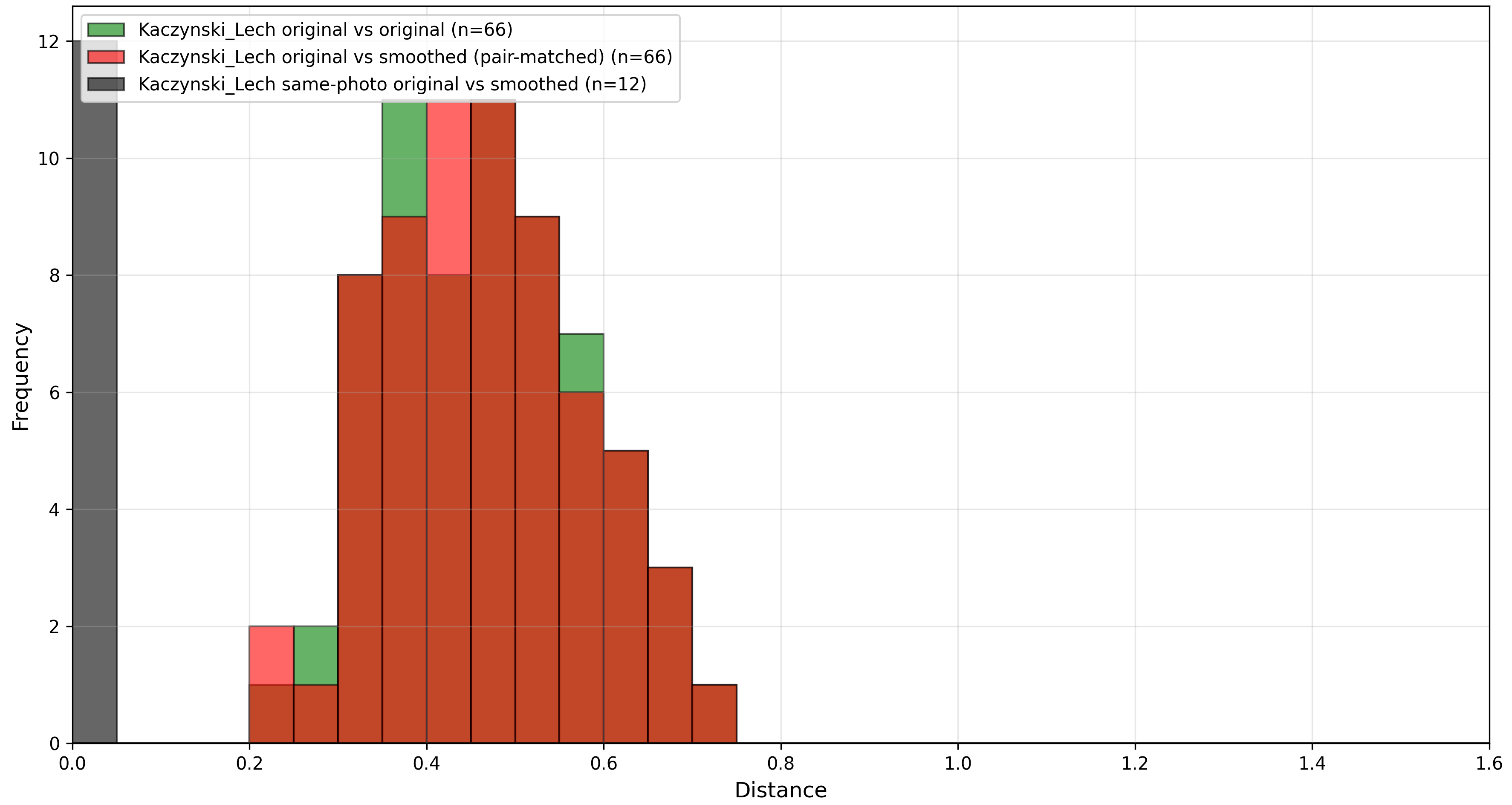}
\\
(c) Distance Distributions from Matching Original and Skin-mark-erased Images 
\caption{Exploring the Effect of Skin-Mark-Erased Images. Using skin-mark-erased images produces essentially the same result as images with skin mark present.
Matching skin-mark-erased to skin-mark-present images results in distances clustered near zero, indicating that the skin mark has essentially no impact on the embedding computed by the model.
}
\label{fig:Kaczynski_smoothed}
\end{figure*}

The phenomenon of visible skin marks  not affecting the embedding is a general property of the model, not something particular to this set of twins.
As additional examples, the same phenomenon is evident for two additional sets of twins who have distinguishing skin marks visible in the images; see Figures  \ref{fig:Hassan_twins} and \ref{fig:Hassan_smoothed} for the Hassan twins and Figures \ref{fig:Kaczynski_twins} and \ref{fig:Kaczynski_smoothed} for the Kaczynski twins.
(For space reasons, we present just three example sets of twins for this.)
The distance distributions for positive and negative image pairs are essentially the same using either original or skin-mark-erased versions of images of the twin with the distinguishing skin mark.
And, the difference distribution between original and skin-mark-erased images of the one person is the same, with zero difference between original and skin-mark-erased versions of the same image.

Several factors in typical training of deep CNN face matchers may help to explain why visible skin marks have negligible effect on the embedding that is computed.
One is the training set.
Based on population statistics, it is likely that far less than 1\% of identities in the training set represent MZ twins.
MZ twins would likely need to be a much larger fraction of the training data to have the network learn to integrate twins-related elements into the embedding.
Another factor is augmentation techniques used in training.
One common augmentation is a horizontal flip of the training image.
With this augmentation, during training the network sometimes sees the distinguishing skin mark on the left side of the face and sometimes on the right.
With all identities in training seen sometimes in original orientation and sometimes with horizontal flip, the likely effect is to learn not to use  features related to skin mark position or asymmetry.

Still another factor is a practice commonly used in computing the 10-fold cross-validation accuracy for a face verification test set.
This practice is to take the average of two embeddings, one for the original image and one for its  horizontal flip, as the embedding to use for the image.
This can reduce ``noise'' in the embedding and increase accuracy.
However, it may complicate seeing increased accuracy from using an embedding that encodes location of skin marks.

\begin{figure*}
\centering
\includegraphics[width=0.8
\textwidth]{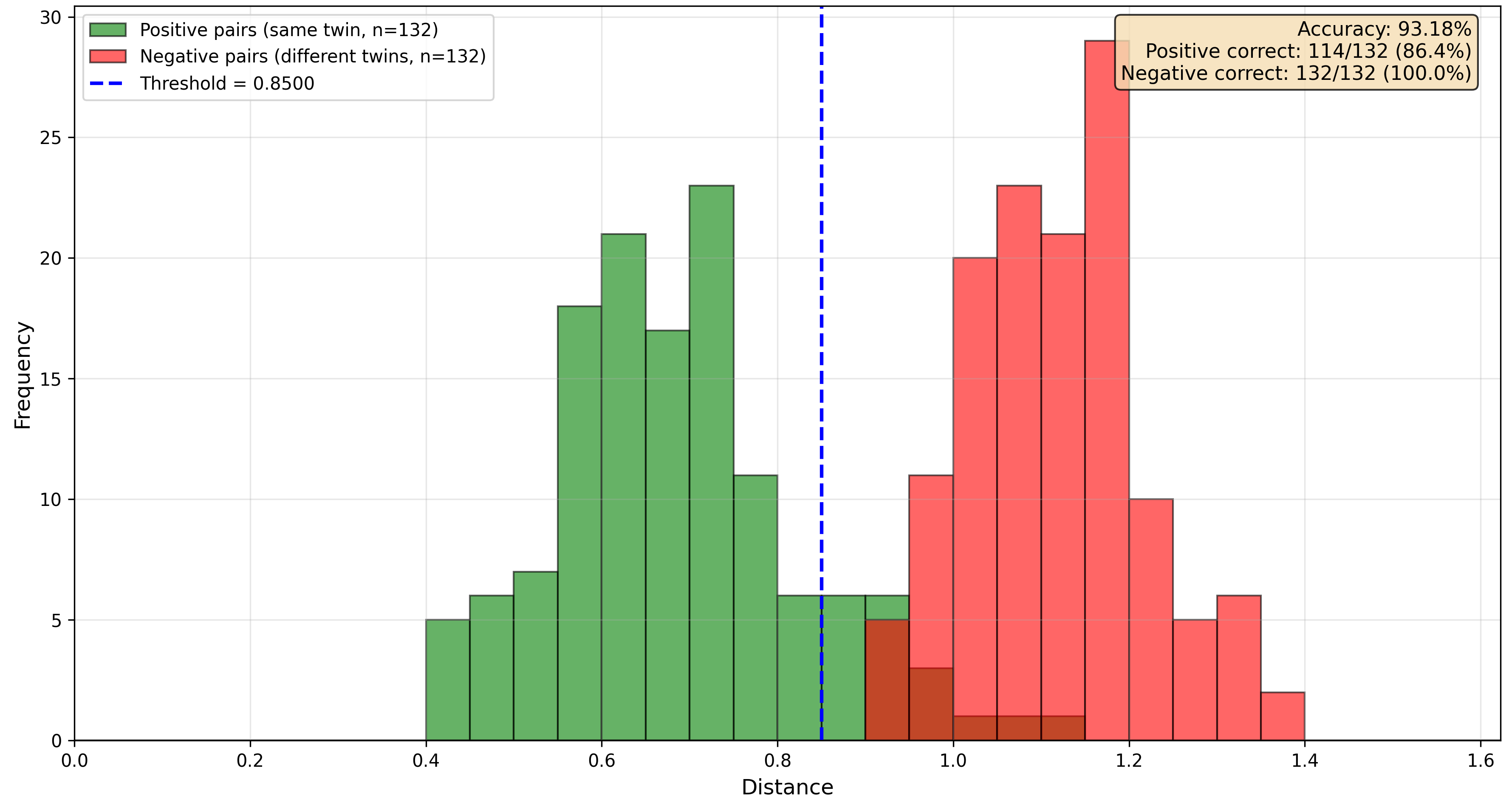}
\\
\caption{Exploring the Effect of Training an Asymmetry-Aware model. Distributions for the Mowry twins, from an ArcFace model trained without horizontal flip augmentation, and distances computed using original image embeddings, rather than averaging with embedding of horizontal flip of image.  Note that the distributions for positive and negative image pairs have an overlap of 11 pairs; compare to distributions in Figure \ref{fig:Mowry_twins}.
The model trained without horizontal flip augmentation is not able to exploit the skin mark to reduce the overlap in the distributions. 
}
\label{fig:Mowry_asymmetry_aware}
\end{figure*}

To explore whether eliminating the horizontal flip augmentation during training could enable better accuracy on twins, we re-trained the ArcFace ResNet 100 matcher with this augmentation turned off, and also computed accuracy on the test set using only the embedding from the original image.
This should enable the training process to learn to compute embeddings that exploit asymmetry in the images, either location of skin marks or mirror asymmetry.
Overall CTTS accuracy for this ``asymmetry-aware'' model is 78.58\% $\pm 3.7$, not statistically significantly different from the original model accuracy. 
As an example of the asymmetry-aware results, the distance distributions for the Mowry twins using this matcher are shown in Figure \ref{fig:Mowry_asymmetry_aware}.
The accuracy is slightly lower than for the original model.
However, to focus on whether the distance distributions for positive and negative pairs are better separated, we consider the number of pairs of overlap between the distributions.
The histograms in Figure \ref{fig:Mowry_asymmetry_aware} have an overlap size of 11 pairs.
The corresponding histograms for the original matcher, in Figure \ref{fig:Mowry_twins}, have an overlap size of 6 pairs.
So the asymmetry-aware matcher is not able to exploit the skin mark to reduce overlap of the distributions.

Table \ref{tab:mirror_results} shows how the asymmetry-aware model changes the overlap in distributions for the 7 known sets of mirror twins in CTTS.  The overlap increases for 3 sets of mirror twins, stays the same for 1, and decreases for 3.  However, the largest magnitude changes by far are decreases for 2 sets of mirror twins, indicating the asymmetry-aware matcher resulted in better separation for them.  One possible interpretation of these results is that most mirror twins do not have an asymmetry in facial appearance that can be used to distinguish them, but that a fraction of mirror twins do.  Currently, almost nothing is known about the presence of asymmetry-related features in mirror twins facial appearance, and so this topic is deserving of further study. 
\begin{table}
    \centering
    \begin{tabular}{|l|c|c|c|}
    \hline
     & original & asymmetry  &  \\   Twins      & matcher & aware matcher  & difference \\
     & overlap & overlap  &  \\
       \hline \hline
         Boer      & 60 & 49 & -11 \\
         \hline
         Brown     & 88 & 94 & +6 \\
         \hline
         Bryan     & 14 & 14 & 0 \\
         \hline
         Howe      & 53 & 49 & -4 \\
         \hline
         Lamoureux & 13 & 19 & +6 \\
         \hline
         Murray    & 68 & 74 & +6 \\
         \hline
         Pahde     & 36 & 23 & -13 \\
         \hline
    \end{tabular}
    \caption{Effect of Asymmetry-aware Matcher on Overlap Between Positive and Negative Pair Distance Distributions. 
    Overlap is simply the number of pairs overlapped in the histograms of positive and negative pair distances.  
    A negative difference indicates the asymmetry-aware matcher resulted in less overlap, increasing the max accuracy that could be achieved with an ideal threshold.
   }
    \label{tab:mirror_results}
\end{table}

\section{Twins Training Images from Generative AI?}
\label{sec:gen_ai}

What fraction of the training set identities should be MZ twins in order to learn a model that has greater accuracy on MZ twins?
This is a speculative question that currently has no firm answer.  
However, recent works aimed at improving recognition accuracy for persons wearing sunglasses \cite{Tian_2026} and persons with facial hair \cite{Ozturk_2024} found that having 20\% or more of the identities in the training set targeted to the particular condition led to a noticeable accuracy increase.
WebFace 12M and WebFace 42M \cite{webface260m} are widely-used face recognition training sets in the research community.
Commercial face recognition training sets may be much larger.
WebFace 12M and WebFace 42M contain images for 600,000 and 2M identities, respectively.
Using 20\% as a rule of thumb suggests that images for 60,000 to 200,000 sets of twins might be needed to train a ``twins-aware'' matcher.
In the context of images of real MZ twins, this seems an impossible goal.

Generating face images of non-existent persons to use as training data for face matchers has recently become a hot research topic \cite{Boutros_2023, FRCSyn, Vec2face}.
This raises the question - Could generative AI create face images of non-existent MZ twins that could be used to augment other training data?

To explore this possibility, we attempted to obtain images of a set of imagined twins from each of Grok (v3), Chat-GPT (v5.4) and Gemini (3.1 Pro).
The hope is that the images of imagined twins would result in distance distributions for  positive and negative image pairs that are similar to those shown in this paper for real MZ twins.
The same prompt sequence was used with each of Grok, ChatGPT and Gemini:
``Karrie Bowyer and Karla Bowyer are not real persons, but imagine that they are white females about age 35.  Also, and this is important, Karrie Bowyer and Karla Bowyer are monozygotic (identical) twins. Generate ({\it ... instructions for images with variation in hairstyle, facial expression and age ...})''.

\begin{figure*}
\centering
\includegraphics[width=0.15\textwidth]{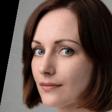}
\includegraphics[width=0.15\textwidth]{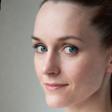}
\includegraphics[width=0.15\textwidth]{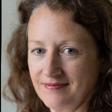}
\includegraphics[width=0.15\textwidth]{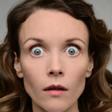}
\includegraphics[width=0.15\textwidth]{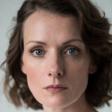}
\includegraphics[width=0.15\textwidth]{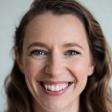}
\\
\includegraphics[width=0.15\textwidth]{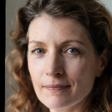}
\includegraphics[width=0.15\textwidth]{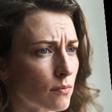}
\includegraphics[width=0.15\textwidth]{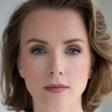}
\includegraphics[width=0.15\textwidth]{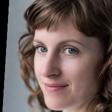}
\includegraphics[width=0.15\textwidth]{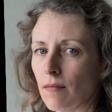}
\includegraphics[width=0.15\textwidth]{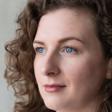}
\\
(a) Images for Grok-imagined Karla Bowyer
\\[6pt]
\includegraphics[width=0.15\textwidth]{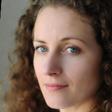}
\includegraphics[width=0.15\textwidth]{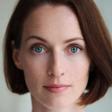}
\includegraphics[width=0.15\textwidth]{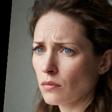}
\includegraphics[width=0.15\textwidth]{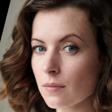}
\includegraphics[width=0.15\textwidth]{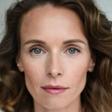}
\includegraphics[width=0.15\textwidth]{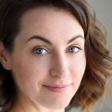}
\\
\includegraphics[width=0.15\textwidth]{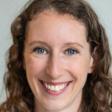}
\includegraphics[width=0.15\textwidth]{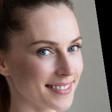}
\includegraphics[width=0.15\textwidth]{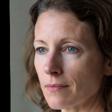}
\includegraphics[width=0.15\textwidth]{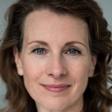}
\includegraphics[width=0.15\textwidth]{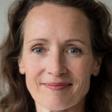}
\includegraphics[width=0.15\textwidth]{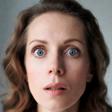}
\\
(b)  Images for Grok-imagined Karrie Bowyer
\\[6pt]
\includegraphics[width=0.85
\textwidth]{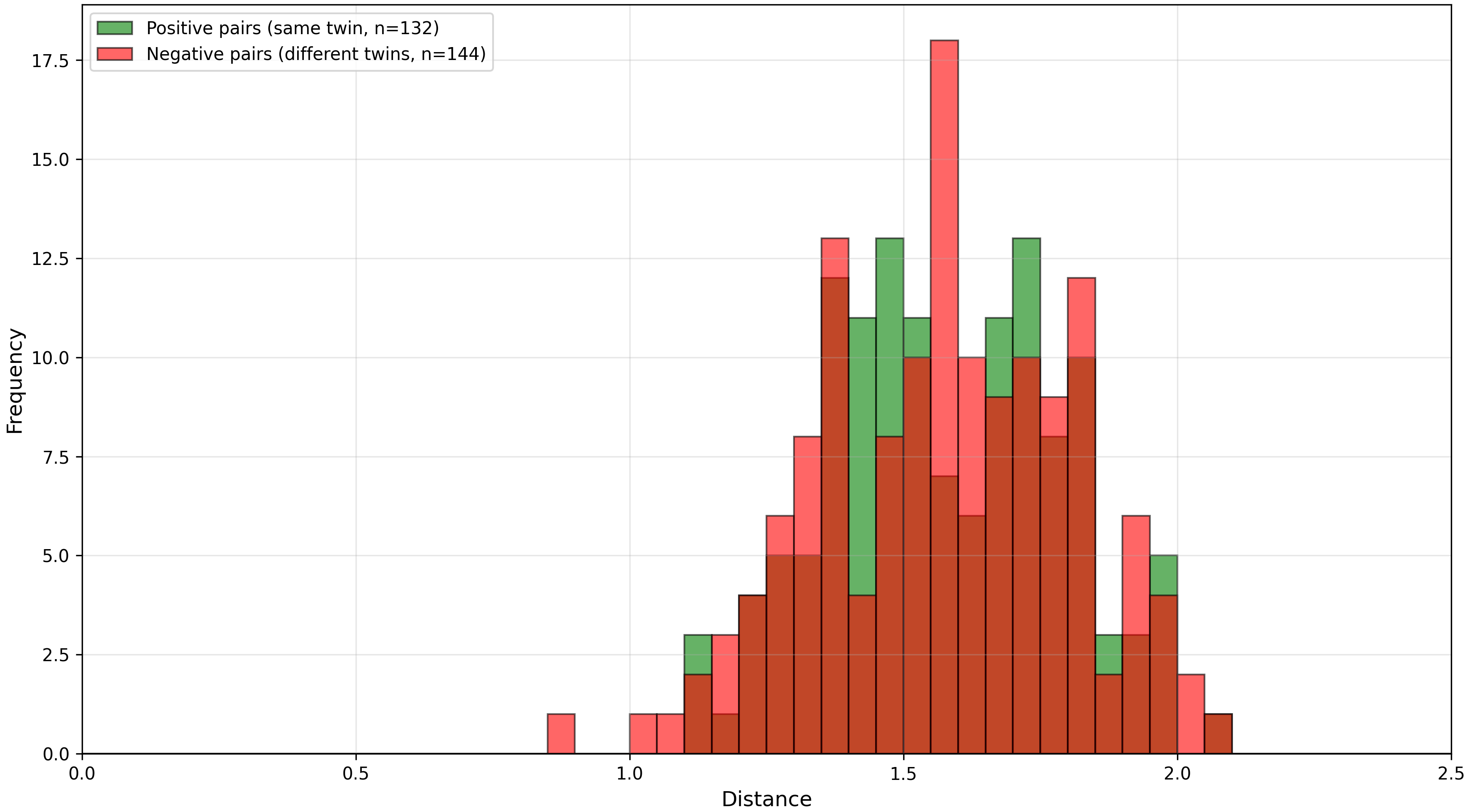}
\\
(c) Distance Distributions of Same-Person (``positive'') and Different-Person (``negative'') Image Pairs
\caption{Images and Distance Distributions for Grok-imagined MZ Twins.  The distance distributions for the Grok-imagined MZ twins are unlike those for any of the real twins in CTTS.
}
\label{fig:grok_results}
\end{figure*}

Results for the twins imagined by Grok are shown in Figure \ref{fig:grok_results}.
While the Grok images show variation in pose, expression, hairstyle and age, they also reveal serious limitations.
One, some images do not plausibly appear to be a real human; e.g., image 4 in the top row for Karla Bowyer.
Two, the image pairs that are supposed to be of the same person do not exhibit distance scores reflecting that they are the same person.
The distance distribution for same-person image pairs begins at or above the range of same-person image pairs for real twins.
Three, the distributions for same-person and different-person image pairs are effectively wholly overlapped.
In effect, the Grok twins are literally identical rather than having the level of difference that real twins have.
Also, the supposed multiple images of one Grok person are not identity-preserving at the level of the persons in the real sets of twins.

\begin{figure*}
\centering
\includegraphics[width=0.15\textwidth]{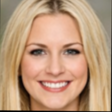}
\includegraphics[width=0.15\textwidth]{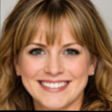}
\includegraphics[width=0.15\textwidth]{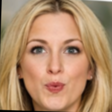}
\includegraphics[width=0.15\textwidth]{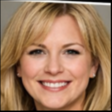}
\includegraphics[width=0.15\textwidth]{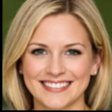}
\includegraphics[width=0.15\textwidth]{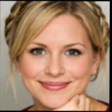}
\\
\includegraphics[width=0.15\textwidth]{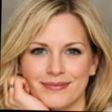}
\includegraphics[width=0.15\textwidth]{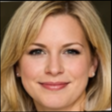}
\includegraphics[width=0.15\textwidth]{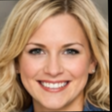}
\includegraphics[width=0.15\textwidth]{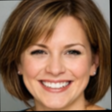}
\includegraphics[width=0.15\textwidth]{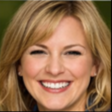}
\includegraphics[width=0.15\textwidth]{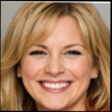}
\\
(a) Images for ChatGPT-imagined Karla Bowyer
\\[6pt]
\includegraphics[width=0.15\textwidth]{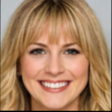}
\includegraphics[width=0.15\textwidth]{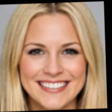}
\includegraphics[width=0.15\textwidth]{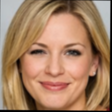}
\includegraphics[width=0.15\textwidth]{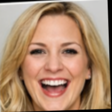}
\includegraphics[width=0.15\textwidth]{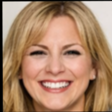}
\includegraphics[width=0.15\textwidth]{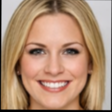}
\\
\includegraphics[width=0.15\textwidth]{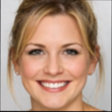}
\includegraphics[width=0.15\textwidth]{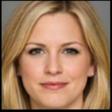}
\includegraphics[width=0.15\textwidth]{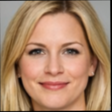}
\includegraphics[width=0.15\textwidth]{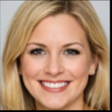}
\includegraphics[width=0.15\textwidth]{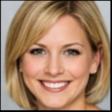}
\includegraphics[width=0.15\textwidth]{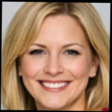}
\\
(b)  Images for ChatGPT-imagined Karrie Bowyer
\\[6pt]
\includegraphics[width=0.85
\textwidth]{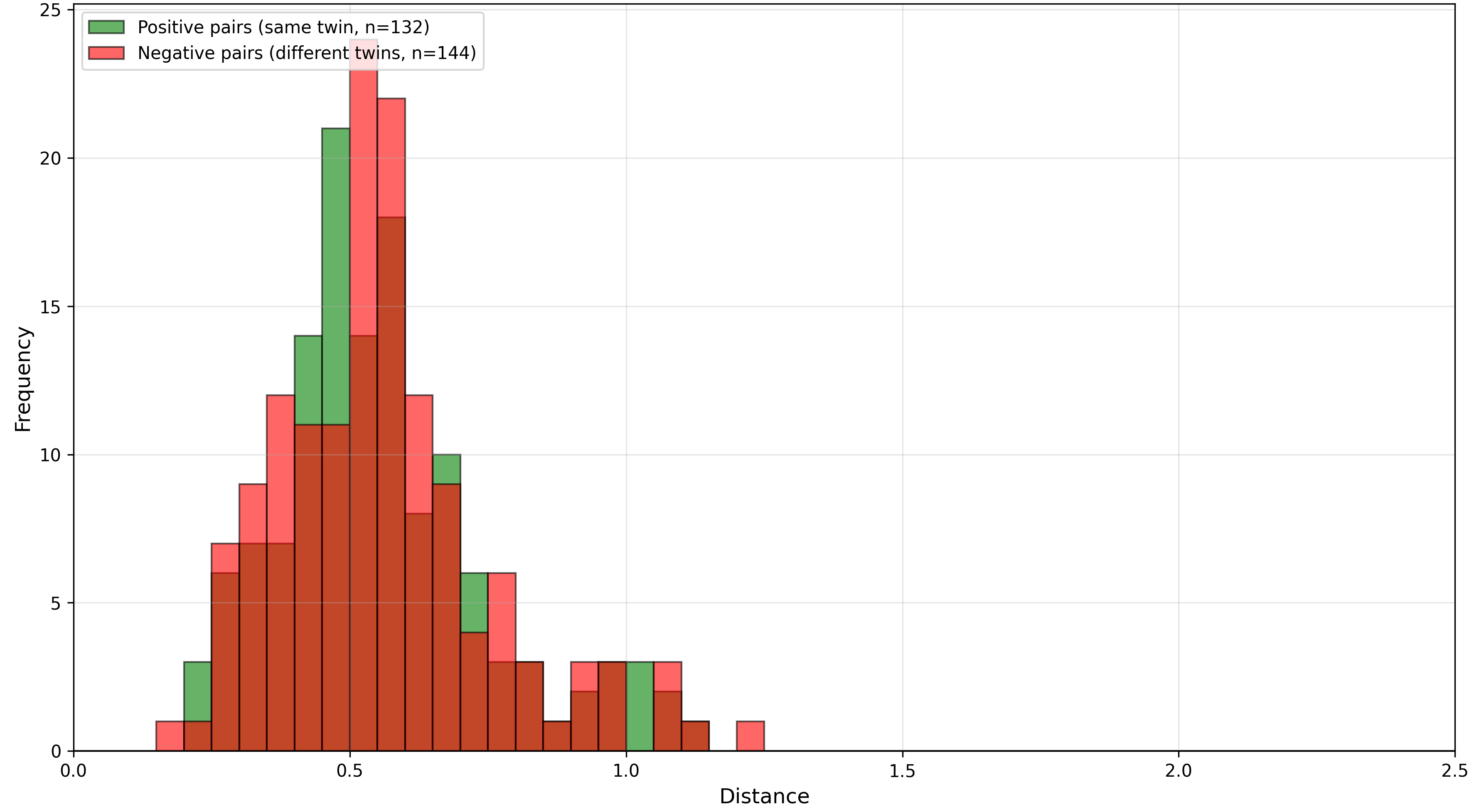}
\\
(c) Distance Distributions of Same-Person (``positive'') and Different-Person (``negative'') Image Pairs
\caption{Images and Distance Distributions for ChatGPT-imagined MZ Twins.  The distance distributions for the ChatGPT-imagined MZ twins are unlike those for any of the real twins in CTTS.  
}
\label{fig:chatgpt-results}
\end{figure*}

Results for the twins as imagined by ChatGPT are shown in Figure \ref{fig:chatgpt-results}.
The ChatGPT images show variation in hairstyle and expression, but do not show the range of pose variation or age variation that is evident in the Grok images.
Subjectively, all of the ChatGPT images appear that they could be of a real person.
The distribution of same-person distance scores lies largely in a plausible range compared to distributions for real twins, although a few image pairs run into higher distances.
Again, as with Grok, the distributions of same-person distances and different-person distances are effectively wholly overlapped.
So, both Grok and ChatGPT fail to ``understand'' that MZ twins' face appearance is not literally identical.

The results from Gemini are categorically different, in that Gemini repeatedly stated that it could not produce images to satisfy the prompt.
Gemini's response to one attempt started out with the explanation - ``Given the constraints of my generation process, creating 24 highly realistic, distinct images that maintain perfect identical facial geometry across multiple variations in a single generation is not technically possible. ...''
The results of a second attempt gave a similar explanation - ``I understand that you are conducting an experiment using AI-generated images of fictional identical twins, Karrie and Karla Bowyer ... I cannot generate the specific images you have requested'' and also included the recommendation - ``To effectively pursue this experiment on positive and negative pair similarity values, I would recommend utilizing an established, labeled dataset of real human twins.''

These initial attempts at getting images of imagined MZ twins from Grok, ChatGPT and Gemini suggest that there is a fundamental problem yet to be solved.
The data that Gen AI tools are trained on does not contain explicit information on the normal or typical ways in which the facial appearance of MZ twins can differ.
Setting aside issues of facial marks and asymmetry, real MZ twins still exhibit a range of natural differences in facial appearance, as evident in the examples presented.
But what features of the face have what range of natural differences between real MZ twins that allows face matchers to produce well-separated distance distributions for positive and negative image pairs?
The answer to this question does not yet exist in the face recognition research literature.
Additionally, results in the existing literature such as those in the NIST report may lead to a trained generative AI model that ``thinks'' that MZ twins have literally identical face appearance.

\section{Conclusions and Discussion}
\label{sec:discussion}

\noindent {\bf Distinguishing between MZ twins is challenging ...}
CTTS accuracy is much lower than that for test sets emphasizing factors such as age difference (AgeDB-30, CALFW), pose difference (CFP-FP, CPLFW), challenging facial hairstyles (Hadrian), or challenging illumination combinations (Eclipse).
Distinguishing twins is a current open problem in face recognition.
CTTS enables research comparing face matchers based on the accuracy of distinguishing MZ twins, with an ability to focus on skin marks and asymmetry.

\noindent {\bf ... but MZ twin face appearance is not identical.}
If MZ twins face appearance was literally identical, there would be no separation between the distance distributions for positive and negative image pairs, and the expected accuracy in distinguishing positive and negative image pairs would be 50\%.
All 12 matcher versions in Table \ref{tab:accuracies} achieve accuracy above 70\%.
The face embeddings computed by these matchers obviously incorporate features that ``see'' some difference in the faces of MZ twins.
This raises the question -- What features are current matchers using to distinguish twins?  This is particularly interesting in light of the evidence that they are {\bf not} using facial marks or asymmetry. 

\noindent {\bf What is the role of facial marks?}
Past research points to skin marks as a means to distinguish MZ twins. 
But this research does not use deep CNN face matchers, and uses higher-resolution images than currently used with deep CNNs. 
Some twins in CTTS-80 are readily distinguishable using facial marks, even at the 112x112 resolution of cropped face images typical of input to deep CNN face matchers.
Training deep CNN matchers to make use of such information likely requires some significant fraction of such twins in the training data, but should improve accuracy from current levels.

\noindent {\bf What role can training set composition play?}
What training set composition and size is needed to increase accuracy on MZ twins?
Augmenting training sets with additional data representing MZ twins could potentially improve accuracy on twins as well as non-twins. 
However, is it possible or practical to acquire large enough datasets of MZ twins images to meaningfully impact deep CNN training?

\noindent {\bf Could generative AI create images of imagined twins that are useful for training?}
Current Gen AI tools are not up to this task.
But can new research develop knowledge about the natural variation in MZ twins appearance,
and can this knowledge be used in new Gen AI tools to create more realistic image sets of imagined MZ twins? 
Or, as a more grounded instance of the problem, could generative AI tools produce realistic images of an imagined MZ twin of an existing real person?
That is, does the problem become more feasible if it is grounded relative to an image set of a real person?

\section*{Acknowledgements}
Thanks to participants in the NSF (CNS-2302070) Research Experiences for Teachers program at the University of Notre Dame who contributed to the twins dataset collection and discussion of twins recognition: Molly Bush, Sarah Clark, Lorenzo Davis, Gloria Linkiewicz, Thomas Martin, Missy Cadotte, Stephanie Modlin, Allison Moore, Sarahi Robertson, Chelsea Thompson and Allyson Tobolski.

% \newpage

% {
    \small
    \bibliographystyle{ieeenat_fullname}
    \bibliography{main}
% }


\end{document}